\documentclass[10pt,journal]{IEEEtran}
\usepackage{cite}
\usepackage{array,multirow,graphicx}
\usepackage{amsmath}
\usepackage{amsthm}
\usepackage{amssymb}
\usepackage{amsfonts}
\usepackage{algorithmicx}
\usepackage{algpseudocode}
\usepackage[caption=false,font=footnotesize,labelfont=sf,textfont=sf]{subfig}
\usepackage{threeparttable}
\usepackage{color}
\usepackage{array}
\usepackage{rotating}
\usepackage{array}
\usepackage{mathrsfs}
\usepackage{titling}
\newtheorem{thm}{Theorem}
\newtheorem{appThm}{Theorem}
\newcolumntype{C}[1]{>{\centering\let\newline\\\arraybackslash\hspace{0pt}}m{#1}}
\newcommand*\rot{\rotatebox{90}}

\begin{document}

\title{Diversifying Support Vector Machines for Boosting using Kernel Perturbation: Applications to Class Imbalance and Small Disjuncts}

\author{Shounak Datta, Sayak Nag, Sankha Subhra Mullick, and Swagatam Das,~\IEEEmembership{Senior Member, IEEE}%
\IEEEcompsocitemizethanks{\IEEEcompsocthanksitem Shounak Datta, Sankha Subhra Mullick, and Swagatam Das are with the Electronics and Communication Sciences Unit, Indian Statistical Institute, Kolkata, India. Sayak Nag is with the Department of Instrumentation and Electronics Engineering, Jadavpur University, Kolkata, India. \protect
\IEEEcompsocthanksitem Corresponding author: Swagatam Das (Email: swagatam.das@isical.ac.in)}}
\date{\vspace{-5ex}}

\maketitle

\begin{abstract}
The diversification (generating slightly varying separating discriminators) of Support Vector Machines (SVMs) for boosting has proven to be a challenge due to the strong learning nature of SVMs. Based on the insight that perturbing the SVM kernel may help in diversifying SVMs, we propose two kernel perturbation based boosting schemes where the kernel is modified in each round so as to increase the resolution of the kernel-induced Reimannian metric in the vicinity of the datapoints misclassified in the previous round. We propose a method for identifying the disjuncts in a dataset, dispelling the dependence on rule-based learning methods for identifying the disjuncts. We also present a new performance measure called Geometric Small Disjunct Index (GSDI) to quantify the performance on small disjuncts for balanced as well as class imbalanced datasets. Experimental comparison with a variety of state-of-the-art algorithms is carried out using the best classifiers of each type selected by a new approach inspired by multi-criteria decision making. The proposed method is found to outperform the contending state-of-the-art methods on different datasets (ranging from mildly imbalanced to highly imbalanced and characterized by varying number of disjuncts) in terms of three different performance indices (including the proposed GSDI).
\end{abstract}


\IEEEpeerreviewmaketitle

\section{Introduction}\label{intro}

\IEEEPARstart{S}{upport} \emph{Vector Machines} (SVMs) \cite{cortes1995support} are a family of popular classifiers having elegant mathematical basis that can be used to model both linear and non-linear (using the kernel trick) decision boundaries. An SVM aims to find the maximum-margin hyperplane
\begin{equation*}
f(\mathbf{x}) = \mathbf{w}^T \mathbf{x} + b,
\end{equation*}
which, for a given training dataset ${P} = \{(\mathbf{x}_{i},{y_{i}})| i \in \{1,\cdots,n\}, y_i \in \{-1,+1\}\}$, is expressed as a linear combination of some of the training points known as the Support Vectors (SVs), i.e.
\begin{equation*}
f(\mathbf{x}) = \sum_{\mathbf{x}_i \in SV} \lambda_i y_i \mathbf{x}_i^T \mathbf{x} + b,
\end{equation*}
where $SV \subseteq P$ is the set of SVs while $\lambda_i$ and $b$ are the classifier parameters. The kernel trick is used to map the data to a higher dimensional feature space in order to facilitate linear separability between classes not linearly separable in the native input space. Due to the use of the kernel trick, the mapping $\Phi(\mathbf{x})$ to the feature space need not be explicitly known. However, the inner product in the feature space is known to be defined by a positive-semidefinite kernel function
\begin{equation*}
K(\mathbf{x}_i,\mathbf{x}) = \Phi(\mathbf{x}_i)^T \Phi(\mathbf{x}).
\end{equation*}
Therefore, the decision boundary in the feature space can be expressed as
\begin{equation*}
f(\mathbf{x}) = \sum_{\mathbf{x}_i \in SV} \lambda_i y_i K(\mathbf{x}_i,\mathbf{x}) + b.
\end{equation*}
While being highly effective for non-overlapping classes, the performance of SVMs suffers in case of overlapping classes, due to the presence of data irregularities such as class imbalance (under-represented classes) \cite{datta2015near, galar2012review, he2009learning} and small disjuncts (under-represented sub-concepts within classes) \cite{weiss2000quantitative, weiss2005mining, garcia2015dissimilarity}. Class imbalanced often results in greater misclassification from the minority class. The presence of small disjuncts, on the other hand, results in low accuracy on the under-represented sub-concepts. In fact, the two problems are closely related as the sub-concepts from the minority class in a class imbalanced problem are often smaller than those of the majority class.

One possible way to deal with such data irregularities is to employ ensemble learning techniques such as \emph{boosting} \cite{freund1995desicion}. Boosting techniques function by learning and ultimately combining a sequence of diverse classifiers such that the latest classifier lays more stress on the correct classification of datapoints misclassified by the ensemble of the previous classifiers. The most popular method to achieve such diversity is to re-sample the misclassified points prior to learning the subsequent classifier. However, due to the stable nature of SVMs, such re-sampling may still result in the generation similar decision boundaries, unless new SVs are created by the re-sampling \cite{rangel2005boosting}). Yet, works like \cite{rangel2005boosting,li2008adaboost,guo2017exclusivity} attest to the fact that proper diversification of SVMs may indeed result in substantial improvement of performance. Li \emph{et al.} \cite{li2008adaboost} proposed an effective way to diversify Radial Basis Function (RBF) kernel based SVMs. The popularity of the RBF kernel (also known
as Gaussian RBF kernel or simply as Gaussian kernel) is due to its ability to map the data to an infinite dimensional feature space, where the pair-wise proximity between points can be controlled by using the parameter $\sigma$. It is defined as
\begin{equation}\label{eqn:rbfK}
K(\mathbf{x}_i,\mathbf{x}_{i'}) = \text{exp} \begin{pmatrix} -\frac{\|\mathbf{x}_i - \mathbf{x}_{i'}\|^{2}}{2\sigma^{2}} \end{pmatrix}.
\end{equation}
References to the kernel function in the rest of this article are about the RBF kernel, unless stated otherwise. The principal idea of \cite{li2008adaboost} was to modify the kernel so that the pair-wise
distances between the points are magnified in each round. However, the method proposed in \cite{li2008adaboost} is too general; hence being unable to handle data irregularities like class imbalance or small disjuncts.

\subsection{Motivation}\label{motiv}

The approach of Li \emph{et al.} \cite{li2008adaboost}, despite being rather general, established that kernel perturbation techniques can be used to diversify SVMs for ensemble learning. However, there exists a need for incorporating selective kernel perturbation to handle data irregularities.

An independent literature exists on the selective perturbation
of kernels (see Section \ref{sec:bg}), based on the initial findings of
Amari and Wu \cite{amari1999improving, wu2002conformal}. The essential idea is to iteratively perturb the kernel so as to accord high resolution (better discriminability) in select neighborhoods of the dataset. A number of such kernel perturbation methods focus on handling class imbalance by selectively applying class-specific perturbation \cite{wu2005kba, williams2005scaling, maratea2011asymmetric, maratea2014adjusted, zhang2014imbalanced}. However, these methods are not equipped to handle local data irregularities such as small disjuncts. Like traditional boosting methods \cite{freund1995desicion}, a mechanism to empower the more challenging datapoints is required to handle this difficulty.

Moreover, there is no guarantee that the performance improvement achieved by the selective kernel perturbation methods will be monotonic over iterations. Therefore, it will be beneficial to use an ensemble of all the SVMs trained on the perturbed kernels obtained in the different iterations (rather than to only use the SVM trained in the final perturbed kernel).

Hence, we are motivated to propose the diversification of SVMs using datapoint-specific kernel perturbation, and to form an ensemble of the SVMs trained in each iteration a la boosting. The principal idea is to accord high resolution around the misclassified datapoints in the subsequent rounds. We note that such an approach not only results in an effective diversification of SVMs, but is also likely to be inherently immune to class imbalance (as the high rate of misclassification from the minority class will result in increased resolution around it). Additionally, the point-specific approach is likely to offer added immunity against small disjuncts by enabling local perturbations.

\subsection{Contribution}\label{contrib}

The principal contributions of this article are as follows:
\begin{enumerate}
\item We propose a point-specific kernel perturbation based scheme for boosting SVMs, which we call Kernel Perturbation based Boosting of SVM (KPBoost-SVM). The proposed scheme (described in Section \ref{methods}) functions by inducing high resolution in the vicinity of the misclassified points for the subsequent round.
\item We also modify the assignment of weights for each of the component classifiers generated by iterative application of the proposed kernel perturbation schemes. The proposed weight generation method accords equal importance to the proper classification of both classes and is therefore expected to be more effective, especially in the presence of data irregularities such as class imbalance and small disjuncts.
\item Since most of the existing performance measures for small disjuncts are dependent on rule-based learning schemes and also do not pay any heed to the class imbalance problem (with which the small disjuncts problem is closely associated) \cite{weiss2000quantitative, carvalho2005evaluating}, we propose a new performance measure called Geometric Small Disjunct Index ($GSDI$) which is capable of measuring the performance on small disjuncts for any classifier (not restricted by the type of classifier) and also does not ignore the presence of class imbalance. It is well-known that the small disjuncts from the minority class(es) are more prone to misclassification compared to those of the majority class(es) \cite{quinlan1991improved}. $GSDI$ can detect such situations as it attains a low value if the small disjuncts from any of the classes suffer a high degree of misclassification, irrespective of the performance on the small disjuncts from the other class(es).
\item In order to use the $GSDI$ for quantifying the performance on small disjuncts, it is essential to identify the disjuncts present in the data. The existing techniques to identify the disjuncts in a dataset are specific to rule-based learning schemes and cannot be applied to other classifiers. Therefore, we propose a classifier-independent algorithm (in Section \ref{indices}), to effectively identify the disjuncts in a dataset.
\item Different performance indices for imbalanced classification often do not agree as to which, out of a given set of classifiers, is the best. In addition, most performance indices do not guarantee the selection of the best trade-off solution between the classes. However, most of the performance indices have the similar underlying motivation to identify the classifier that maximizes all the class-wise classification accuracies. Based on this understanding, we propose a measure inspired by multi-criteria decision to choose the best out of a given set of classifiers, in Section \ref{MCDMchoose}. The best classifiers of each type are chosen by this measure, for comparing the proposed learners with state-of-the-art methods on two-class and multi-class datasets in Section \ref{experiments}.
\end{enumerate}

\section{Background on Selective Kernel Perturbation}\label{sec:bg}

The selective kernel perturbation methods are based on the original findings of Amari and Wu \cite{amari1999improving, wu2002conformal}, who showed that the Reimannian metric component $g_{rs}(\mathbf{x})$ induced in the input space is directly derived from the kernel as
\begin{equation*}
g_{rs}(\mathbf{x}) = \frac{\partial}{\partial x_r} \frac{\partial}{\partial x'_s} K(\mathbf{x'},\mathbf{x})|_{\mathbf{x'} = \mathbf{x}},
\end{equation*}
where $x_r$ and $x_s$ are the $r$-th and $s$-th features of $\mathbf{x}$. They then used this insight to enlarge the spatial resolution around the decision boundary (and diminish it elsewhere) to enhance separability between the classes, using a conformal mapping (i.e. a mapping that preserves the angles between datapoints in the feature space). Then the SVM must be retrained using the modified kernel. The general transformation is of the form
\begin{equation}\label{eqn:modKernel}
K'(\mathbf{x'},\mathbf{x}) = D(\mathbf{x'})K(\mathbf{x'},\mathbf{x})D(\mathbf{x}),
\end{equation}
where $D(\mathbf{x'})$ and $D(\mathbf{x})$ are the non-negative transformation factors at the points $\mathbf{x'}$ and $\mathbf{x}$, respectively (for a proof  that (\ref{eqn:modKernel}) is indeed a conformal mapping, see Corollary 1 of \cite{wu2005kba}). Since the ideal decision boundary is unknown, it was proposed \cite{amari1999improving} to enlarge the resolution around the SVs identified by the initially learned SVM (as SVs are supposed to be close to the decision boundary). The resulting transformation factor around an arbitrary point $\mathbf{x}$ is
\begin{equation*}
D(\mathbf{x}) = \sum_{\mathbf{x}_i \in SV} \lambda_i \times \text{exp} \begin{pmatrix} -\frac{\| \mathbf{x} - \mathbf{x}_i \|^2}{2 \tau^2} \end{pmatrix},
\end{equation*}
where $\lambda_i$ determines the relative importance of each SV while $\tau$ is a tunable parameter which controls the decay of the transformation factors as one moves away from the SVs. Since the SVs are often scattered irregularly around the boundary, $D(\mathbf{x})$ was redefined in \cite{wu2002conformal} as
\begin{equation*}
D(\mathbf{x}) = \sum_{\mathbf{x}_i \in SV} \text{exp} \begin{pmatrix} -\frac{\| \mathbf{x} - \mathbf{x}_i \|^2}{\tau_i^2} \end{pmatrix},
\end{equation*}
where $\tau_i$ is a SV-specific parameter, accounting for the density of the other SVs around the SV in question to compensate for the irregular distribution of SVs. Based on the ground-work laid down in \cite{amari1999improving, wu2002conformal}, Wu and Chang \cite{wu2003adaptive} proposed a modification called Adaptive Conformal Transformation (ACT) to asymmetrically perturb the kernel on the two sides of the decision boundary using
\begin{equation*}
D(\mathbf{x}) = \sum_{\mathbf{x}_i \in SV^{+}} \text{exp} \begin{pmatrix} -\frac{\| \mathbf{x} - \mathbf{x}_i \|^2}{\eta_p \tau_i^2} \end{pmatrix} + \sum_{\mathbf{x}_i \in SV^{-}} \text{exp} \begin{pmatrix} -\frac{\| \mathbf{x} - \mathbf{x}_i \|^2}{\eta_n \tau_i^2} \end{pmatrix},
\end{equation*}
where $SV^{+}$ is the set of positive (minority) SVs, $SV^{-}$ is the set of negative (majority) SVs, while $\eta_p$ and $\eta_n$ are additional tunable parameters which control the asymmetric decay of the transformation factors. Generally, $\eta_p > \eta_n$ is chosen so that the resolution remains higher for the minority class points. Wu and Chang \cite{wu2004aligning, wu2005kba} also proposed the Kernel Boundary Alignment (KBA) technique where the spatial resolution is maximized around an interpolated decision boundary which is placed closer to the majority class to compensate for the over-regularization of the minority class. The transformation factors for KBA are defined as
\begin{equation*}
D(\mathbf{x}) = \frac{1}{|\chi_b^*|} \sum_{\mathbf{x}_b \in \chi_b^*} \text{exp} \begin{pmatrix} - \frac{\|\Phi(\mathbf{x}) - \Phi(\mathbf{x}_b)\|^2}{\tau_b^2} \end{pmatrix},
\end{equation*}
where $\chi_b^{*}$ is the set of interpolated boundary points $\Phi(\mathbf{x}_b)$ and $\tau_b$ is the corresponding parameter accounting for the density of the other interpolated boundary points in the vicinity. Both ACT and KBA are iterative methods where the kernel obtained in each iteration is further perturbed in the subsequent iteration and the SVM is retrained. Williams \emph{et al.} \cite{williams2005scaling} dispelled the unwanted dependence of these techniques on the distribution of SVs by putting forth a different form of the transformation factor viz. 
\begin{equation*}
D(\mathbf{x}) = e^{- k f(\mathbf{x})^2},
\end{equation*}
where $f(\mathbf{x})$ is the output of the initial SVM and $k$ is the decay parameter. This method, known as Kernel Scaling (KS), was extended to the case of class imbalance by Maratea \emph{et al.} \cite{maratea2011asymmetric, maratea2014adjusted}, giving rise to an Asymmetric Kernel Scaling (AKS) method, where different decay parameters $k_{+}$ and $k_{-}$ $(0< k_{+} < k_{-})$ are used for the positive and negative classes, i.e.
\begin{equation*}
D(\mathbf{x}) =
\begin{cases}
e^{- k_{+} f(\mathbf{x})^2} \text{ if } f(\mathbf{x}) >= 0, \\
e^{- k_{-} f(\mathbf{x})^2} \text{ if } f(\mathbf{x}) < 0.
\end{cases}
\end{equation*}
However, the proper values of $k_{+}$ and $k_{-}$ have to be selected through a very costly grid search over $\mathbb{R}^{+} \times \mathbb{R}^{+}$. Moreover, the choice of the decay parameter for an unknown point has to be made based on the output of the initial SVM, which may itself be mis-calibrated due to class imbalance. Furthermore, unlike prior methods, they chose not to retrain the SVM on the perturbed kernel (instead using the already trained SVM with the modified kernel), presumably to cut-down on the computational cost, as the grid search imposes a high computational burden. This essentially reduces AKS to a weighting scheme, not unlike \cite{imam2006z}. Zhang \emph{et al.} \cite{zhang2014imbalanced}, while extending AKS to the multi-class scenario (AKS-$\chi^2$), addressed these issues by proposing to choose the $k_i$ (decay parameter corresponding to the $i$-th class) values based on the $\chi^2$ statistic and to retrain the SVM iteratively, as prevalent in the prior literature.

\section{Kernel Perturbation based Boosting of SVMs}\label{methods}


While kernel perturbation methods like those of \cite{amari1999improving, wu2002conformal, wu2003adaptive, wu2004aligning, wu2005kba, williams2005scaling, maratea2011asymmetric, maratea2014adjusted, zhang2014imbalanced} can accord diversity to SVMs, these are only aimed towards class imbalance and do not address the issue of small disjuncts, which is often responsible for a large fraction of the errors \cite{weiss2000quantitative}. The diversification achieved by these methods is not directed to benefit the difficult points (and thus cannot directly address the small disjunct problem), unlike traditional boosting methods where more stress is laid on the correct classification of difficult points in the subsequent rounds. Hence, to modify the kernel perturbation scheme so that more stress is laid on the misclassified points in the subsequent rounds, we propose KPBoost-SVM. KPBoost-SVM is characterised by a datapoint-specific transformation factor (as opposed to the class-specific transformation factors of the existing methods) of the form
\begin{equation}\label{eqn:Di}
D_t(\mathbf{x}_i)=e^{-k_{i}^t f_{t-1}^{2}(\mathbf{x}_i)},
\end{equation}
where $D_t(\mathbf{x}_i)$ is the transformation factor at the $t$-th round, $k_{i}^t$ serves as the datapoint-specific perturbation parameter at the $t$-th round, and $f_{t-1}(\mathbf{x}_i)$ is the prediction from the $(t-1)$-th round, all corresponding to the training point $\mathbf{x}_i$. Starting with $k^{1}_{i} = 0$ (i.e. $D_1(\mathbf{x}_i)=1)$ $\forall \mathbf{x}_i \in P$, the values of $k_{i}^t$ in each round are updated as
\begin{equation}\label{eqn:perturb}
k_{i}^{t} =
\begin{cases}
k_{i}^{t-1} \hspace{0.72cm} \forall i \notin \Omega, \\
k_{i}^{t-1} + \varsigma \text{ } \forall i \in \Omega,
\end{cases}
\end{equation}
where $\Omega = \{i|h_{t-1}(\mathbf{x}_i) = y\}$ is the set of denominations corresponding to training points correctly classified in the previous round, $h_{t-1}(\mathbf{x}_i) = sign(f_{t-1}(\mathbf{x}_i))$ is the label assigned to $\mathbf{x}_i$ at the $(t-1)$-th round, and $\varsigma > 0$ is a user-specified \emph{kernel perturbation step} used to decrease the resolution around correctly classified training points. In other words, the resolution around the correctly classified datapoints in decreased, thus relatively increasing the resolution around the misclassified points. The following theorem shows that such an update of the kernel perturbation parameters ensures higher resolution for misclassified points close to the decision boundary in the subsequent round.

\begin{thm}
For a datapoint $\mathbf{x}_i \in P$ close to the decision boundary $f_t(\mathbf{x}) = 0$ at the $t$-th round we have $m'_{t+1}(\mathbf{x}_i) > m_{t+1}(\mathbf{x}_i)$, where $m'_{t+1}(\mathbf{x}_i)$ is the ratio of the new resolution at the $(t+1)$-th round and the old resolution at the $t$-th round in the vicinity of $\mathbf{x}_i$ if $\mathbf{x}_i$ were to be misclassified at the $t$-th round and $m_{t+1}(\mathbf{x}_i)$ is the analogous ratio if $\mathbf{x}_i$ were to be correctly classified at the $t$-th round.
\begin{IEEEproof}
See Appendix \ref{append1}.
\end{IEEEproof}
\end{thm}

Starting with $K_{ii'}^{0} = K(\mathbf{x}_i,\mathbf{x}_{i'})$, the Gaussian RBF kernel given by (\ref{eqn:rbfK}), the kernels using which the SVMs are trained for each round are calculated as
\begin{equation}\label{eqn:Kij}
K_{ii'}^{t}=D_t(\mathbf{x}_i)K_{ii'}^{t-1}D_t(\mathbf{x}_{i'}).
\end{equation}
Then, the SVM trained at the $t$-th round can be expressed as
\begin{equation*}
f_t(\mathbf{x}_j) = \sum_{\mathbf{x}_i \in SV_t} \lambda_i^t y_i K_{ij}^t + b^t,
\end{equation*}
where $\mathbf{\Lambda}^t = [\lambda_1^t, \cdots, \lambda_n^t]^{\text{T}}$ ($n$ being the number of training points) and $b^t$ are the learned classifier parameters, and $\mathbf{x}_j$ is a test point which is assigned the label $h_{t}(\mathbf{x}_j) = sign(f_{t}(\mathbf{x}_j))$.

Since, for a class imbalanced classification problem, the minority class datapoints are generally misclassified more than the majority class datapoints, the perturbation scheme of (\ref{eqn:Di}) results in higher spatial resolution around the minority class datapoints; two of the minority training points have high resolution while all majority points have low resolution). Moreover, since most of the misclassification from either of the classes is due to the smaller disjuncts, (\ref{eqn:Di}) also results in high resolution around the small disjuncts, aiding their proper classification.

\begin{figure}[!ht]
\begin{algorithmic}
\small
\State \hrulefill
\State \textbf{Algorithm 1} KPBoost-SVM.
\State \hrulefill
\State \textbf{Input:} Training set ${P} = \{(\mathbf{x}_{i},{y_{i}})| i \in \{1,\cdots,n\}, y_i \in \{-1,+1\}\}$, Test set ${Q} = \{(\mathbf{x}_{j},{y_{j}})| j \in \{1,\cdots,m\}, y_j \in \{-1,+1\}\}$, Number of rounds $T$, Perturbation step $\varsigma$
\State \textbf{Output:} Final classifier $H(\mathbf{x}) = sign(\sum_{t=1}^T \alpha_{t}h_{t}(\mathbf{x}))$
\State \hrulefill

\State \textbf{Training procedure:}
\State Set $f_0(\mathbf{x}_i) = 0 \text{ } \forall \mathbf{x}_i \in P$.
\State Calculate $K_{ii'}^0 = K(\mathbf{x}_i,\mathbf{x}_{i'}) \text{ } \forall \mathbf{x}_i, \mathbf{x}_{i'} \in P$ using (\ref{eqn:rbfK}).
\State Set $k_{i}^1 = 0$ $\forall i \in \{1,\cdots,N\}$.
\For{$t = 1$ to $T$}
    \State Calculate $D_{t}(\mathbf{x}_{i}) = \text{exp}(-k_{i}^{t}f_{t-1}^2(\mathbf{x}_{i}))$.
    \State Update $K_{ii'}^t = D_{t}(\mathbf{x}_i)K_{ii'}^{t-1}D_{t}(\mathbf{x}_{i'})$.
    \State Train an SVM using the perturbed kernel $[K_{ii'}^t]_{N \times N}$.
    \State Find perdicted labels $h_{t}(\mathbf{x}_i)$ using the trained SVM $\forall \mathbf{x}_i \in P$.
    \State Decrease the resolution around the correctly classified datapoints using (\ref{eqn:perturb}).
    \State Calculate $\epsilon_{t} = \sqrt{(1-tpr_t)^{2} + (1-tnr_t)^{2}}$.
    \State Calculate $\alpha_{t} = \frac{1}{2}\log(\frac{(\sqrt 2 - \epsilon_{t})}{\epsilon_{t}})$.
\EndFor
\State $\mathcal{T} = \{t | \epsilon_{t} \leq \min(\epsilon_{1},\frac{1}{\sqrt{2}}) \text{ and } tpr_{t} \geq tpr_{1}\}$.
\If{$\mathcal{T} = \phi$}
    \State $\mathcal{T} = \{t | \epsilon_{t} \leq \frac{1}{\sqrt{2}} \text{ and } tpr_{t} \geq tpr_{1}\}$.
\EndIf
\State $\alpha_t = 0$ $\forall t \notin \mathcal{T}$.\\

\State \textbf{Testing procedure:}
\State Set $D({\mathbf{x}_j}) = 1$ $\forall \mathbf{x}_j \in Q$.
\State Calculate $[K_{ij}^0]_{n \times m}$, where $K_{ij}^0 = K(\mathbf{x}_i,\mathbf{x}_j)$ is given by (\ref{eqn:rbfK}).
\For{$t = 1$ to $T$}
    \State Calculate $K_{ij}^t = D_{t}(\mathbf{x}_i)K_{ij}^{t-1}D(\mathbf{x}_j)$, $\forall \mathbf{x}_i \in P$, $\forall \mathbf{x}_j \in Q$.
    \State Find perdicted labels $h_{t}(\mathbf{x}_j)$ using the SVM trained in round $t$, $\forall \mathbf{x}_j \in Q$.
\EndFor
\State Calculate $H(\mathbf{x}_j) = sign(\sum_{t=1}^T \alpha_{t}h_{t}(\mathbf{x}_j))$ $\forall \mathbf{x}_j \in Q$.

\State \hrulefill
\end{algorithmic}
\end{figure}

\subsection{Choice of weights for the component classifiers}

Training SVMs on the perturbed kernels obtained by $T$ recurring applications of (\ref{eqn:Kij}) results in $T$ different classifiers. These $T$ component classifiers must then be combined by assigning weights $\alpha_t$ based on their individual classification performance on the training set $P$. Hence, the final ensemble classifier is of the form
\begin{equation}\label{eqn:finalH}
H(\mathbf{x}_j) = sign(\sum_{t = 1}^{T} \alpha_{t}h_{t}(\mathbf{x}_j)).
\end{equation}

Traditional boosting techniques like AdaBoost \cite{freund1995desicion} strive to minimize the rate of misclassification
\begin{equation*}
\tilde{\epsilon} = \frac{fp + fn}{tp + fn + tn + fp},
\end{equation*}
where $tp$ and $fn$ are the numbers of training points from the positive class which respectively correctly classified and misclassified by the ensemble classifier, while $tn$ and $fp$ are the corresponding measures for the negative class. However, for a class imbalanced problem, such a measure assigns greater importance to performance on the majority class. Hence, the performance on the two classes must be measured separately and then combined in some meaningful way. Therefore, we instead choose to weigh the component classifiers based on an error metric which measures the distance between the performance of a component classifier and the ideal performance of $(tpr, tnr) = (1, 1)$. The error metric for the $t$-th classifier is defined as
\begin{equation*}
\epsilon_t = \sqrt{(1-tpr_t)^{2} + (1-tnr_t)^{2}},
\end{equation*}
where $tpr_t = \frac{tp_t}{tp_t + fn_t}$ and $tnr_t = \frac{tn_t}{tn_t + fp_t}$ are the true positive and true negative rates for the $t$-th component classifier. Hence, along the lines of AdaBoost, the weight $\alpha_{t}$ of the $t$-th classifier is calculated as
\begin{equation*}
\alpha_t = \frac{1}{2} \log \begin{pmatrix} \frac{\sqrt{2} - \epsilon_{t} }{\epsilon_{t}} \end{pmatrix},
\end{equation*}
$\sqrt{2}$ being the maximum possible value of $\epsilon_t$. Now, to further weed-out component classifiers having poor performance, we find the set
\begin{equation*}
\mathcal{T} = \{t | \epsilon_{t} \leq \min(\epsilon_{1},\frac{1}{\sqrt{2}}) \text{ and } tpr_{t} \geq tpr_{1}\}
\end{equation*}
of the rounds having an error metric value less than half of its maximum possible value ($\epsilon_t \leq \frac{1}{\sqrt{2}}$) and having performance (in terms of both $\epsilon_t$ as well as $tpr_t$) not worse than the initial SVM trained using the unperturbed kernel (first round). If $\mathcal{T}$ is found to be empty, it is relaxed as
\begin{equation*}
\mathcal{T} = \{t | \epsilon_{t} \leq \frac{1}{\sqrt{2}} \text{ and } tpr_{t} \geq tpr_{1}\},
\end{equation*}
to allow the inclusion of rounds with greater $\epsilon_t$ yet having better or equal $tpr_t$, compared to the initial SVM. This relaxed set is never empty and at least contains the first round. All classifiers not belonging to the set $\mathcal{T}$ are assigned zero weights, i.e. $\alpha_t = 0 \text{ } \forall t \notin \mathcal{T}$.

\subsection{Classifying a test point}

When an unknown test point $\mathbf{x}_j$ is to be classified, the sequence of transformation factors $D_t(\mathbf{x}_j)$ is unknown. Therefore, to allow maximum resolution to the new point, we assign $D(\mathbf{x}_j)=1$\footnotemark. Then the initial kernel values $K_{ij}^{0} = K(\mathbf{x}_i,\mathbf{x}_j) \text{ } \forall \mathbf{x}_i \in P$ are calculated using (\ref{eqn:rbfK}), following which the kernel values for the test point in question can be calculated for each round using the expression
\begin{equation*}
K_{ij}^t = D_{t}(\mathbf{x}_i)K_{ij}^{t-1}D(\mathbf{x}_j), \text{ } \forall \mathbf{x}_i \in P.
\end{equation*}
The final prediction $H(\mathbf{x}_j)$ for the test point $\mathbf{x}_j$ is then obtained using (\ref{eqn:finalH}).

\footnotetext{The reader may note that $D(\mathbf{x}_j)=1$ is unknown for a test point. Discussion on a possible solution to this issue can be found in the supplementary document.}

\subsection{The complete KPBoost-SVM algorithm}

The complete KPBoost-SVM algorithm (including the testing procedure) is presented in Algorithm 1. The following theorems deal with the computational complexity of the KPBoost-SVM method.

\begin{thm}
The training computational complexity for KPBoost-SVM is $O(n^2 T)$.
\begin{IEEEproof}
See Appendix \ref{append2}.
\end{IEEEproof}
\end{thm}

\begin{thm}
The testing computational complexity for KPBoost-SVM is $O(nmT)$.
\begin{IEEEproof}
See Appendix \ref{append2}.
\end{IEEEproof}
\end{thm}

\section{Experiments}\label{experiments}

\subsection{Indices for evaluation of classification performance}\label{indices}

In this section, we present the results of experiments conducted using the proposed method on various two-class and multi-class datasets plagued by data irregularities like class imbalance and small disjuncts up to varying degrees. The experimental results are reported by us in terms of three different performance indices, namely \emph{Geometric mean (Gmean)}, \emph{Area Under the Curve (AUC)}, and \emph{Geometric Small Disjunct Index (GSDI)} (which we first propose herein). The three indices are briefly explained in the following:


\subsubsection{Gmean} Kubat and Matwin \cite{kubat1997addressing} proposed the $Gmean$ index which has become one of the most popular performance measures for imbalanced data classification. Since most classification algorithms are unlikely to perform equally well on all the classes in an imbalanced dataset, $Gmean$ reports the geometric mean of the classification accuracies achieved on the individual classes. This is a rather unforgiving index as poor performance on any one of the classes yields a low $Gmean$.

\subsubsection{AUC} The $AUC$ index \cite{maloof2003learning} is the area under the receiver operating characteristic curve which is popularly used to measure the performance of classifiers on imbalanced datasets. For a given two-class classifier, $AUC = \frac{1 + tpr - fpr}{2}$, where $fpr = \frac{fp}{fp + tn}$. It has also been extended for multi-class problems in \cite{hand2001simple}.

\subsubsection{GSDI} As the concept of small disjuncts first appeared in relation to decision trees and rule-based learning, the existing indices for measuring the performance on small disjuncts are tailored for such classifiers \cite{weiss2000quantitative, carvalho2005evaluating}. Moreover, these indices are not standalone and must be combined with other indices like classification accuracy in order to draw meaningful inferences. Furthermore, these indices pay no heed to the data imbalance problem, which often occurs along with the problem of small disjuncts. Therefore, inspired by the $Gmean$ index, we propose $GSDI$ which not only measures the performance on small disjuncts but also accounts for the data imbalance in a dataset. For a $C$-class problem, the $GSDI$ value is calculated as
\begin{equation*}
GSDI = \left(\prod_{c=1}^{C} \frac{\sum_{i=1}^{\delta_c} \text{exp}(-|\Delta_i^c|) \times \frac{{m_c^i}'}{m_c^i} }{\sum_{i=1}^{\delta_c} \text{exp}(-|\Delta_i^c|)} \right)^\frac{1}{C},
\end{equation*}
where $\delta_c$ is the number of disjuncts within the $c$-th class while $\Delta_i^c$ is the $i$-th disjunct within the $c$-th class containing $m_c^i$ test points, ${m_c^i}'$ out of which are correctly classified. Like the $Gmean$ index, $GSDI$ also calculates the geometric mean of the individual class accuracies but assigns high weightage to the accuracies on the smaller disjuncts within classes. Thus, poor performance on the small disjuncts from any one class yields a low $GSDI$ value, irrespective of the performance on the small disjuncts from the other classes.

\subsubsection*{Identifying the disjuncts in a dataset for evaluating GSDI}
To be able to apply the proposed $GSDI$ to measure the performance on any dataset, it is important to identify the disjuncts present in a dataset. However, to the best of our knowledge, there exists no technique to identify the disjuncts present in a dataset, outside of the rule-based learning literature where the rules or nodes accounting for a small number of training points are considered to correspond to small disjuncts. Therefore, assuming that disjuncts correspond to subclusters within the classes, we propose an algorithm to identify the disjuncts present in a dataset $X = P \cup Q$, consisting of $C$ classes denoted as $\mathcal{C}_c$ $(c = 1,\cdots,C)$ containing both training and testing points, i.e. $\mathcal{C}_c = \mathcal{C}_c^P \cap \mathcal{C}_c^Q$. The proposed algorithm runs Breadth First Search (BFS) separately on each of the $C$ classes, assuming each datapoint $\mathbf{x} \in \mathcal{C}_c$ to be linked to $\kappa_c$ $(\kappa_c \leq |\mathcal{C}_c|)$ of its nearest neighbors within the same class. The essential idea behind this approach is that the BFS will identify the distinct connected-components (i.e. disjuncts) within each class.

\begin{figure}[!ht]
\begin{algorithmic}
\small
\State \hrulefill
\State \textbf{Algorithm 2} Identifying the disjuncts in a dataset.
\State \hrulefill
\State \textbf{Input:} Dataset $X = \bigcup_{c=1}^{C} \mathcal{C}_c$, parameter $\kappa \in \mathbb{Z}^{+}$.
\State \textbf{Output:} Sets of disjuncts $\Delta^{c}_{i}, \forall c = 1, 2, \cdots ,C, i = 1, 2, \cdots ,\delta_{c}$.
\State \hrulefill

\State Initialise queue $\mathcal{Q}$.
\ForAll{$c \in \{1, 2, \cdots, C\}$}
    \State $\kappa_c = min(\kappa_c, |\mathcal{C}_c|)$.
    \State Set $\delta_{c} = 1$.
    \State $\mathcal{M} = \phi$.
    \While{($\mathcal{C}_{c} \setminus \mathcal{M}) \neq \phi$}
        \State Find $\mathbf{x} \in (\mathcal{C}_{c} \setminus \mathcal{M})$.
        \State Set $\Delta_{\delta_{c}}^{c} = \{\mathbf{x}\}$.
        \State Update $\mathcal{M} = \mathcal{M} \cup \mathbf{x}$.
        \State $insertQueue(\mathcal{Q}, \mathbf{x})$.
        \While{$\mathcal{Q} \neq \phi$}
            \State $\mathbf{u} \leftarrow deleteQueue(\mathcal{Q})$.
            \State $\Gamma(\mathbf{u}) \leftarrow nearestNeighbors(\kappa_c,\mathbf{u},X) \cap \mathcal{C}_{c}$.
            \ForAll{$\mathbf{v} \in (\Gamma(\mathbf{u}) \setminus (\mathcal{M} \cup \mathcal{Q}))$}
                \State Update $\Delta_{\delta_{c}}^{c} = \Delta_{\delta_{c}}^{c} \cup \mathbf{v}$.
                \State Update $\mathcal{M} = \mathcal{M} \cup \mathbf{v}$.
                \State $insertQueue(\mathcal{Q}, \mathbf{v})$.
            \EndFor
        \EndWhile
        \State $\delta_{c} = \delta_{c} + 1$.
    \EndWhile
\EndFor

\State \hrulefill
\end{algorithmic}
\end{figure}

\begin{figure}[!th]
\begin{center}

\subfloat[Iris:12vs3]{\includegraphics[width=0.23\textwidth, height=1.1in]{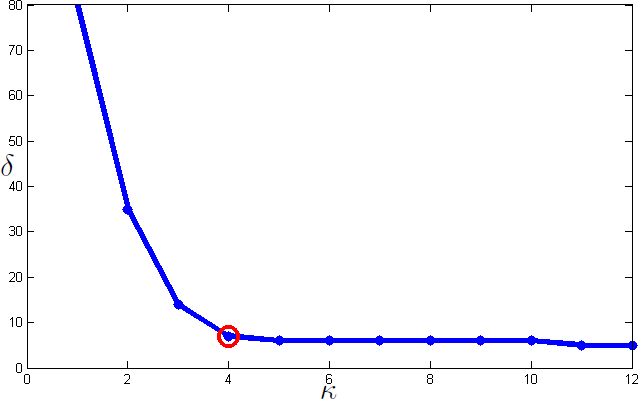}
\label{fig:iris}}
\hfill
\subfloat[Balance]{\includegraphics[width=0.23\textwidth, height=1.1in]{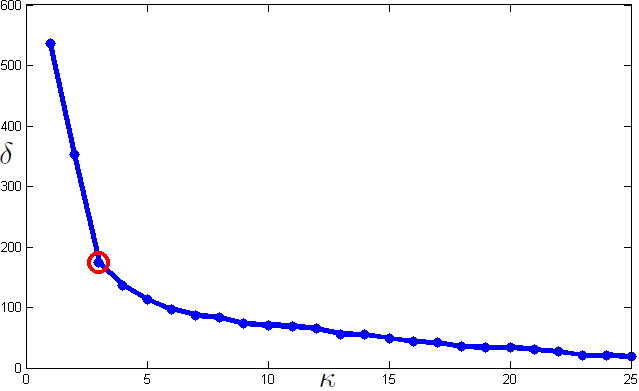}
\label{fig:balance}}

\caption{$\kappa-\delta$ curves for Iris:12vs3 and Balance datasets with the knee-points.}
\label{fig:kappaDelta}

\end{center}
\end{figure}

The entire process is detailed as Algorithm 2 which requires the input of a user-specified parameter $\kappa$. Since $\kappa$ determines the size of the connected neighborhoods, a small value of $\kappa$ results in a large number of disjuncts while a large value divides the dataset into very few disjuncts. Therefore, we obtain a downward-sloping $\kappa-\delta$ ($\delta=\sum_{c=1}^{C}{\delta_{c}}$ being the total number of disjuncts) curve for any dataset by varying the value of $\kappa$ in the range $[1,\sqrt{N}]$ ($N = |P \cap Q|$). We choose the disjuncts corresponding to the ``knee point'' \cite{Kaplan2012} of the $\kappa-\delta$ curve as being the true disjuncts of a dataset. These are used to evaluate the GSDI. The $\kappa-\delta$ curve as well as the chosen knee-point are illustrated for two of the datasets used in our experiments in Figure \ref{fig:kappaDelta}.
The following theorem deals with the comutational complexity of the disjunct identifying method presented in Algorithm 2.

\begin{thm}
The computational complexity for a single run of the method presented in Algorithm 2 is $O(N + \kappa N)$.
\begin{IEEEproof}
See Appendix \ref{append3}.
\end{IEEEproof}
\end{thm}

\subsection{Description of the datasets}
We use 62 two-class and 21 multi-class imbalanced datasets (respectively 10 and 9 of which are high-dimensional, i.e. having more than 500 dimensions) for our experiments. These datasets are derived from various standard datasets \cite{lecun1998gradient, Lichman2013, Fabris2016, triguero2017}. The datasets are chosen so as to range from mild to very high degree of imbalance (Imbalance Ratio, i.e., the ratio of the number of representatives from the largest to that of the smallest class, varying from 1.47 to 853) and also to be composed of a varying number of disjuncts (ranging from 11 disjuncts to 657 disjuncts), as identified by Algorithm 2. The details of all the datasets can be found in the supplement to this article. 
\subsection{Contending algorithms and parameter settings}

The various state-of-the-art methods chosen for comparison with the proposed methods are as follows:

\subsubsection{SVM}
SVM is chosen as a baseline algorithm so that the extent of improvement achieved by the kernel perturbation methods can be estimated. The regularization parameter $C_r$ is varied in the set $\{100,1000\}$. As datsets of varying dimensionality are used in our experiments, the parameter $\sigma$ for the RBF kernel is varied in $\{0.01, 0.02, \cdots, 0.1, 0.2, \cdots, 1, 2,$ $\cdots, 10, 20, \cdots, 100, 200\}$.

\subsubsection{RUSBoost} Galar \emph{et al.} \cite{galar2012review} conducted extensive experiments and found the RUSBoost algorithm \cite{seiffert2010rusboost} to be the most effective boosting scheme for imbalanced datasets. Hence, we include the RUSBoost algorithm (with C4.5 \cite{quinlan2014c4} as the base learner) for comparison with the proposed method. The parameters for the underlying C4.5 are chosen as per \cite{galar2012review}. The percentage of minority points sampled is varied as 35\%, 50\% and 65\% for two-class cases as recommended in \cite{seiffert2010rusboost}. For multi-class datasets, equal number of points are sampled from each class. RUSBoost is trained for 10 rounds.

\subsubsection{AdaBoost-SVM and AdaBoost-MLP}
The prior work in the field of kernel perturbation based boosting of SVMs is AdaBoost-SVM \cite{li2008adaboost}. It is included as a contending algorithm to illustrate the increased efficacy achieved by the proposed techniques. The parameter $\sigma_{step}$ is varied in the set $\{1, 2, 3\}$ and 40 rounds are used, as per \cite{li2008adaboost}. We also compare the proposed method against a boosted variant of Multi-Layer Perceptron (MLP) called AdaBoost-MLP \cite{schwenk2000boosting}, which is a popular ensemble learning method. The number of hidden nodes for AdaBoost-MLP is varied in the set $\{10, 15, 20\}$.

\subsubsection{C4.5/IB1 and D-SVM}
Carvalho and Freitas \cite{carvalho2005evaluating}, based on their experimental findings, concluded that the C4.5/IB1 algorithm \cite{ting1994} is the most effective method for handling small disjuncts, making it a natural candidate for comparison. The threshold on the node size $\varrho$ below which a node is considered to correspond to a small disjunct is varied in $\{3, 5, 10, 15\}$ \cite{carvalho2005evaluating}. Garc{\'\i}a \emph{et al.} \cite{garcia2015dissimilarity}, on the other hand, observed that a D-SVM, i.e. SVM trained on the distance space (where each datapoint is represented as the vector of its distances from the training points) exhibits resilience to small disjuncts as well as noise. Therefore, D-SVM is also chosen as a contender, for being a recent SVM variant with immunity to small disjuncts. D-SVM is tested on the linear kernel as well as the RBF kernel with $\sigma \in \mathcal{S} = \{\frac{\sigma^*}{2}, \frac{3\sigma^*}{4}, \sigma^*, \frac{5\sigma^*}{4}, \frac{3\sigma^*}{2}\}$ ($C_r=1$ being used as per \cite{garcia2015dissimilarity}), where $\sigma^*$ is the best $\sigma$ observed for the baseline SVM on the dataset in question.

\subsubsection{KBA, ACT, AKS and AKS-$\chi^2$}
We also compare the proposed method with the existing kernel perturbation methods viz. KBA, ACT, AKS and AKS-$\chi^2$ which are tailored for handling class imbalanced classification problems. At most 10 iterations are used for these methods, except AKS-$\chi^2$ for which it is varied in $\{2, 3, 5\}$ due to the tendency to easily overfit. The other parameters (if any) for KBA, ACT and AKS-$\chi^2$ are chosen as per the respective papers along with $C_r \in \{100,1000\}$ and $\sigma \in \mathcal{S}$. For AKS, $k_{+}$ and $k_{-}$ are both varied in the set $S = \{1 \times 10^{-4}, 2 \times 10^{-4}, \cdots, 9 \times 10^{-4}, 1 \times 10^{-3}, 2 \times 10^{-3}, \cdots, 9 \times 10^{-3}, 1 \times 10^{-2}, 2 \times 10^{-2}, \cdots, 9 \times 10^{-2}, 0.1, 0.2, \cdots, 1\}$, also with $\sigma \in \mathcal{S}$ and $C_r \in \{100,1000\}$.

\subsubsection{KPBoost-SVM}
10 rounds of KPBoost-SVM are run with the perturbation step $\varsigma$ being varied as elements in $S$ while $Cr \in \{100,1000\}$ is used.

\subsection{Multi-criteria based selection of the best classifier}\label{MCDMchoose}

While there exist many indices to measure the performance of classifiers on imbalanced datasets, these do not necessarily agree upon a single classifier as being the best among a given group of classifiers. For example, let us consider four classifiers having performance as shown below:

\begin{footnotesize}
\begin{center}
\smallskip
\begin{tabular*}{0.35\textwidth}{c|cccc}
\hline
Classifier & $tpr$ & $tnr$ & $Gmean$ & $AUC$ \\
\hline
1 & 0.400 & 0.800 & 0.5657 & 0.6000 \\
2 & 0.670 & 0.670 & 0.6700 & 0.6700 \\
3 & 0.875 & 0.515 & \textbf{0.6712} & 0.6950 \\
4 & 0.900 & 0.500 & 0.6708 & \textbf{0.7000} \\
\hline
\multicolumn{5}{l}{Best values shown in \textbf{boldface}.}
\end{tabular*}
\end{center}
\end{footnotesize}

We find that $Gmean$ and $AUC$ select different classifiers. Moreover, both indices fail to identify classifier 2 which offers the best trade-off between the classes. Thus, being inspired by a measure prevalent in the field of multi-criteria decision making \cite{koski1984}, we propose the following measure to identify the best trade-off classifier in a set $\mathcal{H}$:
\begin{equation*}
\mu(h_i) = \sum_{c=1}^{C} \frac{( \frac{m'_{c,i}}{m_{c}} - \underset{j \in \mathcal{H}}{\min} \frac{m'_{c,j}}{m_{c}} )}{ ( \underset{j \in \mathcal{H}}{\max} \frac{m'_{c,j}}{m_{c}} - \underset{j \in \mathcal{H}}{\min} \frac{m'_{c,j}}{m_{c}} )},
\end{equation*}
where $m_{c}$ is the number of test points from class $c$, $m'_{c,i}$ of which are correctly classified by the classifier $h_i$. The classifier offering the best trade-off between the classes is expected to have a higher value of $\mu$ compared to the other classifiers in $\mathcal{H}$. We first use 10-fold cross-validation on the datasets (for datasets having less than 10 minority class points, cross-validation is carried out on random subsamples drawn so as to conserve the degree of imbalance) to find the average performances. The best classifier of each type is then selected as the one achieving greatest $\mu$ based on these average performances. Identical partitionings are used for corresponding runs of all the contenders on each dataset.

\subsection{Experimental results for two-class datasets}\label{expTwoClass}

\begin{table*}
\begin{center}
\scriptsize
\caption{Performances on two-class datasets in terms of Average Rank (AR) and Wilcoxon's Signed Rank (WSR) hypotheses.}
\label{ResTwo}
\begin{tabular}{c|cc|cc|cc|cc|cc|cc} \hline
& \multicolumn{6}{c|}{Two-class Low-Dimensional Datasets} & \multicolumn{6}{c}{Two-class High-Dimensional Datasets} \\ \cline{2-13}
Algorithm & \multicolumn{2}{c|}{$Gmean$} & \multicolumn{2}{c|}{$AUC$} & \multicolumn{2}{c|}{$GSDI$} & \multicolumn{2}{c|}{$Gmean$} & \multicolumn{2}{c|}{$AUC$} & \multicolumn{2}{c}{$GSDI$} \\ \cline{2-13}
& AR & WSR & AR & WSR & AR & WSR & AR & WSR & AR & WSR & AR & WSR \\ \hline
SVM	&	4.24	&	$H_{1}$	&	4.24	&	$H_{1}$	&	4.26	&	$H_{1}$	&	4.95	&	$H_{1}$	&	4.80	&	$H_{1}$	&	4.75	&	$H_{1}$	\\
AdaBoost-SVM	&	8.99	&	$H_{1}$	&	8.89	&	$H_{1}$	&	8.79	&	$H_{1}$	&	8.80	&	$H_{1}$	&	8.65	&	$H_{1}$	&	8.55	&	$H_{1}$	\\
C4.5/IB1	&	8.60	&	$H_{1}$	&	8.57	&	$H_{1}$	&	8.01	&	$H_{1}$	&	7.35	&	$H_{1}$	&	7.70	&	$H_{1}$	&	5.90	&	$H_{1}$	\\
D-SVM	&	6.60	&	$H_{1}$	&	6.56	&	$H_{1}$	&	6.15	&	$H_{1}$	&	5.55	&	$H_{1}$	&	5.30	&	$H_{1}$	&	4.85	&	$H_{1}$	\\
AKS	&	3.15	&	$H_{1}$	&	2.98	&	$H_{1}$	&	3.21	&	$H_{0}$	&	3.05	&	$H_{0}$	&	3.05	&	$H_{0}$	&	3.05	&	$H_{0}$	\\
AKS-$\chi^{2}$	&	5.96	&	$H_{1}$	&	6.02	&	$H_{1}$	&	5.49	&	$H_{1}$	&	8.45	&	$H_{1}$	&	8.45	&	$H_{1}$	&	7.55	&	$H_{1}$	\\
RUSBoost	&	6.90	&	$H_{1}$	&	7.19	&	$H_{1}$	&	6.35	&	$H_{1}$	&	7.55	&	$H_{1}$	&	7.70	&	$H_{1}$	&	6.55	&	$H_{1}$	\\
AdaBoost-MLP	&	5.15	&	$H_{1}$	&	5.07	&	$H_{1}$	&	8.14	&	$H_{1}$	&	3.20	&	$H_{0}$	&	3.45	&	$H_{0}$	&	8.60	&	$H_{1}$	\\
KBA	&	7.01	&	$H_{1}$	&	7.29	&	$H_{1}$	&	6.38	&	$H_{1}$	&	6.95	&	$H_{1}$	&	6.75	&	$H_{1}$	&	6.85	&	$H_{1}$	\\
ACT	&	7.32	&	$H_{1}$	&	7.12	&	$H_{1}$	&	6.79	&	$H_{1}$	&	8.20	&	$H_{1}$	&	8.20	&	$H_{1}$	&	7.20	&	$H_{1}$	\\ 
KPBoost-SVM	&	\textbf{2.07}	&	CN &	\textbf{2.07}	&	CN	&	\textbf{2.43}	&	CN	&	\textbf{1.95}	&	CN	&	\textbf{1.95}	&	CN	&	\textbf{2.15}	&	CN	\\
\hline
\multicolumn{13}{l}{Best values shown in \textbf{boldface}.}\\
\multicolumn{13}{l}{CN: Control for Wilcoxon's signed rank test.}\\
\multicolumn{13}{l}{$H_0$: The contender is statistically similar to the control.}\\
\multicolumn{13}{l}{$H_1$: The contender is significantly different compared to the control.}
\end{tabular}
\end{center}
\end{table*}

The results achieved by the selected classifiers over each of the two-class datasets are summarized in Table \ref{ResTwo}, in terms of the average rankings obtained as per the average $Gmean$, $AUC$ and $GSDI$ values achieved by 10-fold cross-validation. The detailed results can be found in the supplementary document. Additionally, the hypotheses obtained by comparing the states-of-the-art with the proposed method using Wilcoxon's signed rank test \cite{wilcoxon1945individual} at $5\%$ level of significance are also reported. The results for low and high dimensional datasets are reported separately. Interestingly, the status quo is similar for all the performance indices over both types of datasets and indicates that the proposed method is generally better at handling both class imbalance as well as small disjuncts, compared to the other contenders. 

It is also interesting to observe that AKS overall performs better than the other state-of-the-art algorithms and performs statistically at par with the proposed KPOBoost-SVM on the high dimensional datasets. Such performance of AKS, despite not retraining the classifier on the modified kernel, can be understood by thinking of AKS as being akin to a class-wise weighting scheme, as already observed in Section \ref{intro}. The exhaustive search for the kernel perturbation parameters $(k_{+},k_{-})$ over $S \times S$ essentially enables AKS to find the best relative weights between the classes. The fact that the proposed method performs slightly better than AKS (which employs exhaustive search over two parameters) as well as AdaBoost-MLP (which can generate a diverse set of component classifiers) attests to the effectiveness of our proposal.

\subsection{Experimental results for multi-class datasets}

\begin{table*}
\begin{center}
\scriptsize
\caption{Performances on multi-class datasets in terms of Average Rank (AR) and Wilcoxon's Signed Rank (WSR) hypotheses}
\label{ResMulti}
\begin{tabular}{c|cc|cc|cc|cc|cc|cc} \hline
& \multicolumn{6}{c|}{Multi-class Low-Dimensional Datasets} & \multicolumn{6}{c}{Multi-class High-Dimensional Datasets} \\ \cline{2-13}
Algorithm & \multicolumn{2}{c|}{$Gmean$} & \multicolumn{2}{c|}{$AUC$} & \multicolumn{2}{c|}{$GSDI$} & \multicolumn{2}{c|}{$Gmean$} & \multicolumn{2}{c|}{$AUC$} & \multicolumn{2}{c}{$GSDI$} \\ \cline{2-13}
& AR & WSR & AR & WSR & AR & WSR & AR & WSR & AR & WSR & AR & WSR \\ \hline
SVM-OVO	&	2.50	&	$H_{0}$	&	2.21	&	$H_{0}$	&	2.67	&	$H_{0}$	&	3.39	&	$H_{1}$	&	3.22	&	$H_{1}$	&	3.61	&	$H_{1}$	\\
AKS-$\chi^{2}$	&	3.38	&	$H_{1}$	&	3.50	&	$H_{1}$	&	3.25	&	$H_{1}$	&	3.44	&	$H_{0}$	&	3.56	&	$H_{1}$	&	3.44	&	$H_{0}$	\\
RUSBoost	&	3.58	&	$H_{1}$	&	4.00	&	$H_{1}$	&	3.96	&	$H_{0}$	&	3.94	&	$H_{1}$	&	4.67	&	$H_{1}$	&	3.83	&	$H_{1}$	\\
AdaBoost-MLP	&	3.63	&	$H_{1}$	&	3.46	&	$H_{1}$	&	3.13	&	$H_{0}$	&	2.28	&	$H_{0}$	&	2.33	&	$H_{0}$	&	2.17	&	$H_{0}$	\\
KPBoost-SVM-OVO	&	\textbf{1.92}	&	CN	&	\textbf{1.83}	&	CN	&	\textbf{2.00}	&	CN	&	\textbf{1.94}	&	CN	&	\textbf{1.22}	&	CN	&	\textbf{1.94}	&	CN	\\
\hline		
SVM-OVA	&	2.88	&	$H_{1}$	&	2.92	&	$H_{1}$	&	3.00	&	$H_{1}$	&	2.83	&	$H_{1}$	&	2.78	&	$H_{1}$	&	3.06	&	$H_{1}$	\\
AKS-$\chi^{2}$	&	3.13	&	$H_{1}$	&	3.13	&	$H_{0}$	&	3.21	&	$H_{0}$	&	3.67	&	$H_{1}$	&	3.56	&	$H_{1}$	&	3.89	&	$H_{1}$	\\
RUSBoost	&	3.58	&	$H_{1}$	&	3.79	&	$H_{1}$	&	3.96	&	$H_{1}$	&	4.33	&	$H_{1}$	&	4.78	&	$H_{1}$	&	4.33	&	$H_{1}$	\\
AdaBoost-MLP	&	3.38	&	$H_{0}$	&	3.17	&	$H_{0}$	&	2.75	&	$H_{0}$	&	2.83	&	$H_{1}$	&	2.67	&	$H_{1}$	&	2.28	&	$H_{0}$	\\ KPBoost-SVM-OVA	&	\textbf{2.04}	&	CN	&	\textbf{2.00}	&	CN	&	\textbf{2.08}	&	CN	&	\textbf{1.33}	&	CN	&	\textbf{1.22}	&	CN	&	\textbf{1.44}	&	CN	\\
\hline
\multicolumn{13}{l}{Best values shown in \textbf{boldface}.}\\
\multicolumn{13}{l}{CN: Control for Wilcoxon's signed rank test.}\\
\multicolumn{13}{l}{$H_0$: The contender is statistically similar to the control.}\\
\multicolumn{13}{l}{$H_1$: The contender is significantly different compared to the control.}
\end{tabular}
\end{center}
\end{table*}

We further illustrate the effectiveness of the proposed methods on multi-class datasets. The results are compared with that of SVM (serving as the baseline) alongside that of AdaBoost-MLP (because of its competitive performance on the two-class datasets) as well as those of RUSBoost and AKS-$\chi^2$ (as these two methods are the only imbalance handling methods that are directly applicable to multi-class datasets) in Table \ref{ResMulti}. One-Versus-One (OVO) and One-Versus-All (OVA) decompositions are used to apply SVM, KPBoost-SVM and KPBoostROI-SVM to the multi-class scenario. It must be noted that the same value of the parameter $\varsigma$ is used for each of the binary classifiers trained within the OVO and OVA frameworks. Similar to Table \ref{ResTwo}, the results are reported in terms of average rankings and Wilcoxon's signed rank test hypotheses, calculated using results from 10-fold cross-validation. The detailed results are presented in the supplement to this article. 

A perusal of the average rankings in Table \ref{ResMulti} shows that both the OVO as well as OVA variants of KPBoost-SVM perform consistently better than or similar to the states-of-the-art on all the performance indices. This indicates that the ability of KPBoost-SVM to handle class imbalance as well as small disjuncts remains undiminished for multi-class datasets as well. An interesting observation one can make from Table \ref{ResMulti} is that the performance of AdaBoost-MLP is statistically comparable to that of the OVA variant of the proposed method on the low-dimensional datasets, but is significantly different than that of the OVO variant on the same datasets. This indicates that KPBoost-SVM-OVO performs better than KPBoost-SVM-OVA on the low dimensional datasets. Indeed, the efficacy of the OVO scheme on low dimensional datasets in also apparent from the good performance of the OVO variant of SVM, which performs statistically at par with the KPBoost-SVM-OVO. Similarly, the KPBoost-SVM-OVA seems to be more effective than its OVO counterpart on the high dimensional datasets. 

\section{Conclusions}\label{conclu}

We propose the KPBoost-SVM method to diversify SVMs for boosting to lend immunity to data irregularities like class imbalance and small disjuncts. The proposed method lends diversity to the SVMs trained in each round by perturbing the kernel. Th proposed method differs fundamentally from the existing kernel perturbation methods in that it neither perturbs the entire kernel-induced space (unlike \cite{li2008adaboost}), nor is the perturbation only applied around the boundary region (unlike \cite{amari1999improving, wu2002conformal, williams2005scaling}), nor is it class-specific (unlike \cite{wu2003adaptive, wu2004aligning, wu2005kba, maratea2011asymmetric, maratea2014adjusted, zhang2014imbalanced}). The proposed kernel perturbation scheme is more akin to the traditional boosting methods in that point-specific perturbation is applied to the kernel in each round to lay more stress on the correct classification of the difficult points. This is done by increasing the resolution of the kernel-induced Reimannian metric around such points. The proposed KPBoost-SVM is expected to compensate for class imbalance by lending more resolution to the points from the minority class as most of the misclassification is likely to be from the minority class. On the other hand, it is also expected to improve the performance on small disjuncts as most of the misclassification is likely to be concentrated at the smaller disjuncts from each class. Experimental comparison with a variety of state-of-the-art methods (on a large number of two-class as well as multi-class datasets plagued by data irregularities) shows that KPBoost-SVM performs competitively against the state-of-the-art.

\begin{appendices}

\section{On the increase in resolution around misclassified points close to the decision boundary}\label{append1}

\begin{appThm}
For a datapoint $\mathbf{x}_i \in P$ close to the decision boundary $f_t(\mathbf{x}) = 0$ at the $t$-th round we have $m'_{t+1}(\mathbf{x}_i) > m_{t+1}(\mathbf{x}_i)$, where $m'_{t+1}(\mathbf{x}_i)$ is the ratio of the new resolution at the $(t+1)$-th round and the old resolution at the $t$-th round in the vicinity of $\mathbf{x}_i$ if $\mathbf{x}_i$ were to be misclassified at the $t$-th round and $m_{t+1}(\mathbf{x}_i)$ is the analogous ratio if $\mathbf{x}_i$ were to be correctly classified at the $t$-th round.
\begin{IEEEproof}
From Theorem 2 of \cite{amari1999improving} we know that for the RBF kernel, the modified Reimannian metric component $\tilde{g'}_{rs}(\mathbf{x})$ when $\mathbf{x}_i$ is misclassified at the $t$-th round is
\begin{equation*}
\tilde{g'}_{rs}(\mathbf{x}) = D'^2_{t+1}(\mathbf{x}_i) g_{rs}(\mathbf{x}) + \frac{\partial D'^2_{t+1}(\mathbf{x}_i)}{\partial x_r} \frac{\partial D'^2_{t+1}(\mathbf{x}_i)}{\partial x_s},
\end{equation*}
while the corresponding component $\tilde{g}_{rs}(\mathbf{x})$ if $\mathbf{x}_i$ is correctly classified at the $t$-th round is
\begin{equation*}
\tilde{g}_{rs}(\mathbf{x}) = D^2_{t+1}(\mathbf{x}_i) g_{rs}(\mathbf{x}) + \frac{\partial D^2_{t+1}(\mathbf{x}_i)}{\partial x_r} \frac{\partial D^2_{t+1}(\mathbf{x}_i)}{\partial x_s},
\end{equation*}
where $D'_{t+1}(\mathbf{x}_i)$ and $D_{t+1}(\mathbf{x}_i)$ are the transformation factors accorded to $\mathbf{x}_i$ when it is misclassified and correctly classified, respectively.
Now, since $\mathbf{x}_i$ is close to the boundary $f_t(\mathbf{x}) = 0$ where the transformation factors attain maximum, we can consider
\begin{equation*}
\begin{aligned}
& \tilde{g'}_{rs}(\mathbf{x}) = D'^2_{t+1}(\mathbf{x}_i) g_{rs}(\mathbf{x}), \text{ and} \\
& \tilde{g}_{rs}(\mathbf{x}) = D^2_{t+1}(\mathbf{x}_i) g_{rs}(\mathbf{x}).
\end{aligned}
\end{equation*}
Therefore, we have
\begin{equation*}
\begin{aligned}
m'_{t+1}(\mathbf{x}_i) & = \sqrt{\frac{det|\tilde{g'}_{rs}(\mathbf{x})|}{det|g_{rs}(\mathbf{x})|}} \\
& = D'^d_{t+1}(\mathbf{x}_i) = \text{exp} (-k^t_i f^2_t(\mathbf{x}_i)), \text{ and} \\
m_{t+1}(\mathbf{x}_i) & = \sqrt{\frac{det|\tilde{g}_{rs}(\mathbf{x})|}{det|g_{rs}(\mathbf{x})|}} \\
& = D^d_{t+1}(\mathbf{x}_i) = \text{exp} (-k^t_i f^2_t(\mathbf{x}_i)) \text{exp} (-\varsigma f^2_t(\mathbf{x}_i)),
\end{aligned}
\end{equation*}
where $d$ is the number of features of $\mathbf{x}_i$.
Since $\varsigma > 0$, we get $m'_{t+1}(\mathbf{x}_i) > m_{t+1}(\mathbf{x}_i)$.
This completes the proof.
\end{IEEEproof}
\end{appThm}

\section{On the computational complexity of KPBoost-SVM}\label{append2}

\begin{appThm}
The training computational complexity for KPBoost-SVM is $O(n^2 T)$.
\begin{IEEEproof}
The computational complexity for training and SVM is known to be $O(n^2)$ \cite{cao2006approximate}. Computing the $n$ transformation factors requires $O(n)$ time. Therefore, a single training round of KPBoost-SVM requires $O(n^2)$ time. Hence, $T$ rounds of training require $O(n^2 T)$ time.
\end{IEEEproof}
\end{appThm}

\begin{appThm}
The testing computational complexity for KPBoost-SVM is $O(nmT)$.
\begin{IEEEproof}
The testing complexity for a single point for an SVM is $O(n)$. Consequently, the complexity for testing a single point using the $T$ SVMs in the KPBoost-SVM ensemble is $O(nT)$. As the number of testing points is $m$, the total testing complexity is $O(nmT)$.
\end{IEEEproof}
\end{appThm}

\section{On the computational complexity of Algorithm 2}\label{append3}

\begin{appThm}
The computational complexity for a single run of the method presented in Algorithm 2 is $O(N + \kappa N)$.
\begin{IEEEproof}
Since the graph formed by assuming each datapoint to be connected to its $\kappa_c$ neighbors from the same class has $N$ vertices and at most $\kappa N$ links, the complexity for a single run of Algorithm 2 is $O(N+\kappa N)$.
\end{IEEEproof}
\end{appThm}

\end{appendices}

\bibliographystyle{IEEEtran}
\bibliography{diverseSVMrefs}

\begin{IEEEbiography}[{\includegraphics[width=1in,height=1.25in,clip,keepaspectratio]{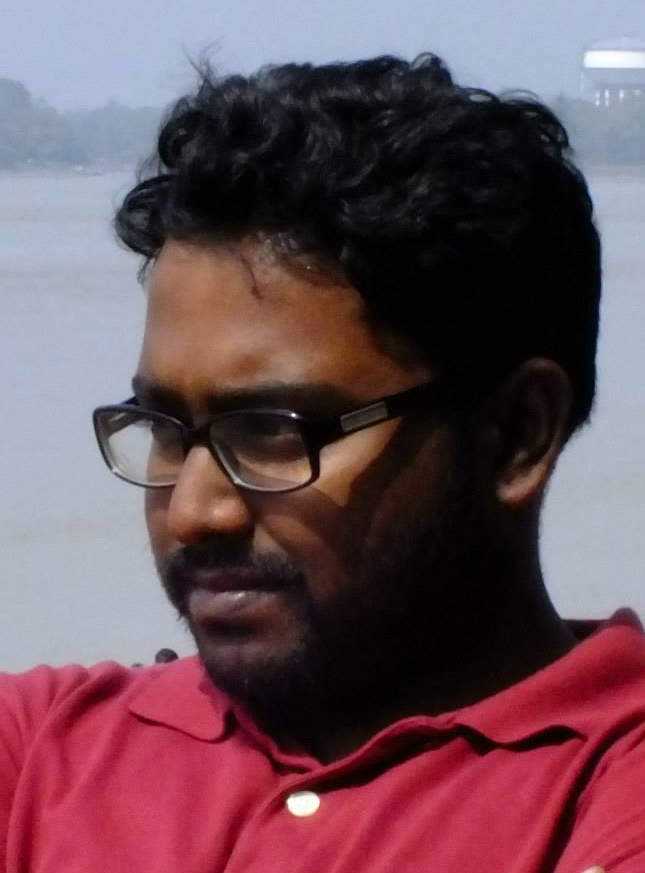}}]{Shounak Datta} received his B.Tech. degree in Electronics and Communication Engineering from the West Bengal University of Technology, Kolkata, India in 2011, and M.E. in Electronics and Telecommunication Engineering from the Jadavpur University, Kolkata, India in 2013. He is currently pursuing a Ph.D. in Computer Science from the Indian Statistical Institute, Kolkata, India. His research interests include imbalanced dataset classification, learning with missing features, multi-objective optimization in machine learning, etc.
\end{IEEEbiography}

\begin{IEEEbiography}[{\includegraphics[width=1in,height=1.1in,clip,keepaspectratio]{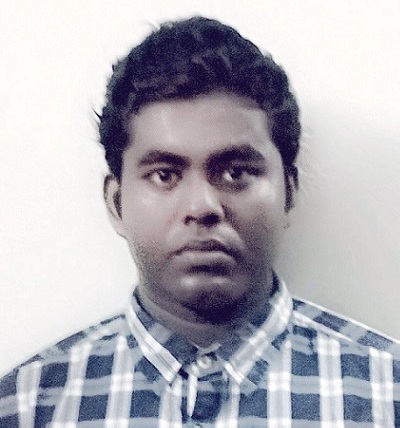}}]{Sayak Nag} is currently pursuing his B.E. in Instrumentation and Electronics Engineering from the Jadavpur University, Kolkata, India. His research interests include ensembles classifiers, support vector machines, neural networks, multi-objective optimization, and machine learning in general.
\end{IEEEbiography}

\begin{IEEEbiography}[{\includegraphics[width=1in,height=1.25in,clip,keepaspectratio]{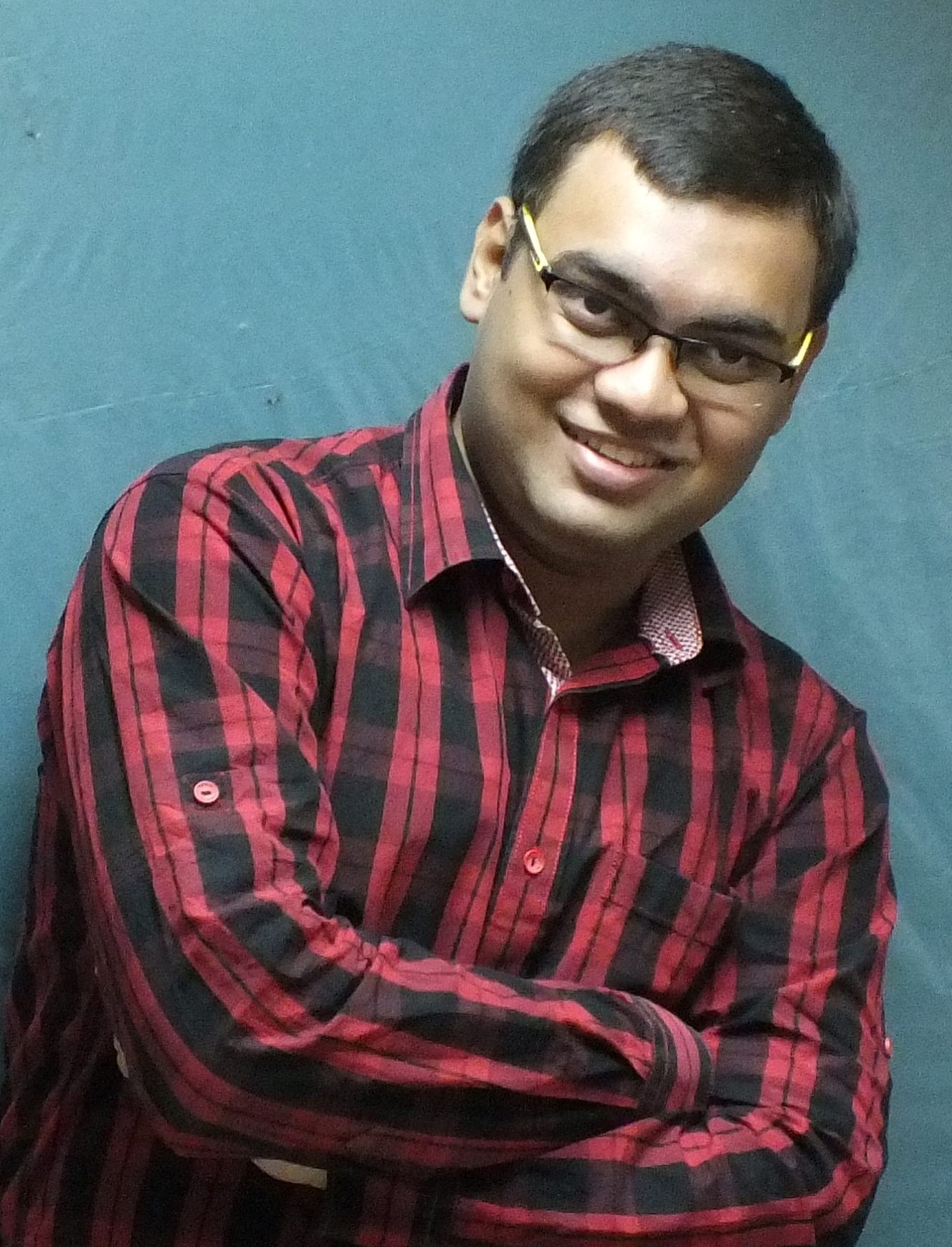}}]{Sankha Subhra Mullick} received his B.Tech in Computer Science and Engineering in 2014 from West Bengal University of Technology. He completed his M.Tech in Computer Science from Indian Statistical Institute in 2014. He is currently pursuing a Ph.D. in Computer Science from Indian Statistical Institute. His research interest include classification and clustering in presence of data anomalies, multi-label learning, optimization techniques and bioinformatics.
\end{IEEEbiography}

\begin{IEEEbiography}[{\includegraphics[width=1in,height=1.25in,clip,keepaspectratio]{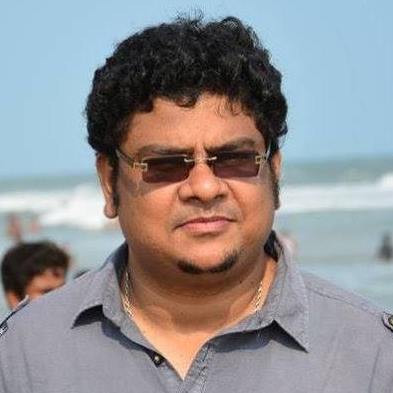}}]{Swagatam Das} is currently serving as a faculty member at the Electronics and Communication Sciences Unit, Indian Statistical Institute, Kolkata, India. He has published one research monograph, one edited volume, and more than 200 research articles in peer-reviewed journals and international conferences. He is the founding co-editor-in-chief of Swarm and Evolutionary Computation, an international journal from Elsevier. Dr. Das has 13,000+ Google Scholar citations and an H-index of 54 till date.
\end{IEEEbiography}

\title{Diversifying Support Vector Machines for Boosting using Kernel Perturbation: Applications to Class Imbalance and Small Disjuncts \\ -Supplementary Material}

\author{Shounak Datta, Sayak Nag, Sankha Subhra Mullick, and Swagatam Das,~\IEEEmembership{Senior Member, IEEE}}
\maketitle 
\IEEEpeerreviewmaketitle

\setcounter{section}{0}

\section{Region Of Influence based KPBoost-SVM}\label{kpboostROI}

\begin{figure}[!th]
\vspace{-3mm}
\begin{algorithmic}
\small
\State \hrulefill
\State \textbf{Algorithm S1} KPBoostROI-SVM.
\State \hrulefill
\State \textbf{Input:} Training set ${P} = \{(\mathbf{x}_{i},{y_{i}})| i \in \{1,\cdots,n\}, y_i \in \{-1,+1\}\}$, Test set ${Q} = \{(\mathbf{x}_{j},{y_{j}})| j \in \{1,\cdots,m\}, y_j \in \{-1,+1\}\}$, Number of rounds $T$, Perturbation step $\varsigma$, ROI scaling parameter $\vartheta \in (0,1]$
\State \textbf{Output:} Final classifier $H(\mathbf{x}) = sign(\sum_{t=1}^T \alpha_{t}h_{t}(\mathbf{x}))$
\State \hrulefill

\State \textbf{Training procedure:}
\State Set $f_0(\mathbf{x}_i) = 0 \text{ } \forall \mathbf{x}_i \in P$.
\State Calculate $K_{ii'}^0 = K(\mathbf{x}_i,\mathbf{x}_{i'}) \text{ } \forall \mathbf{x}_i, \mathbf{x}_{i'} \in P$ using RBF kernel.
\State Set $k_{i}^1 = 0$ $\forall i \in \{1,\cdots,N\}$.
\State Calculate $ROI^{+}$ as the average nearest neighbor distance among the training points from the positive class.
\State Calculate $ROI^{-}$ as the average nearest neighbor distance among the training points from the negative class.
\For{$t = 1$ to $T$}
    \State Calculate $D_{t}(\mathbf{x}_{i}) = \text{exp}(-k_{i}^{t}f_{t-1}^2(\mathbf{x}_{i}))$.
    \State Update $K_{ii'}^t = D_{t}(\mathbf{x}_i)K_{ii'}^{t-1}D_{t}(\mathbf{x}_{i'})$.
    \State Train an SVM using the perturbed kernel $[K_{ii'}^t]_{N \times N}$.
    \State Find perdicted labels $h_{t}(\mathbf{x}_i)$ using the trained SVM $\forall \mathbf{x}_i \in P$.
    \State Decrease the resolution around the correctly classified datapoints which do not belong within the ROIs of any of the misclassified datapoints.
    \State Calculate $\epsilon_{t} = \sqrt{(1-tpr_t)^{2} + (1-tnr_t)^{2}}$.
    \State Calculate $\alpha_{t} = \frac{1}{2}\log(\frac{(\sqrt 2 - \epsilon_{t})}{\epsilon_{t}})$.
\EndFor
\State $\mathcal{T} = \{t | \epsilon_{t} \leq min(\epsilon_{1},\frac{1}{\sqrt{2}}) \text{ and } tpr_{t} \geq tpr_{1}\}$.
\If{$\mathcal{T} = \phi$}
    \State $\mathcal{T} = \{t | \epsilon_{t} \leq \frac{1}{\sqrt{2}} \text{ and } tpr_{t} \geq tpr_{1}\}$.
\EndIf
\State $\alpha_t = 0$ $\forall$ $t \notin \mathcal{T}$.\\

\State \textbf{Testing procedure:}
\State Calculate $[K_{ij}^0]_{n \times m}$, where $K_{ij}^0 = K(\mathbf{x}_i,\mathbf{x}_j)$ is given by RBF kernel.
\State Set $f_0(\mathbf{x}_j) = 0$ for each test datapoint $\mathbf{x}_j \in Q$.
\For{$t = 1$ to $T$}
    \State $\Gamma_j \leftarrow nearestNeighborID(1,\mathbf{x}_j,P)$, $\forall \mathbf{x}_j \in Q$.
    \State $D_{t}(\mathbf{x}_{j}) = \text{exp}(-k_{\Gamma_j}^{t}f_{t-1}^2(\mathbf{x}_{j}))$, $\forall \mathbf{x}_j \in Q$.
    \State Calculate $K_{ij}^t = D_{t}(\mathbf{x}_i)K_{ij}^{t-1}D_{t}(\mathbf{x}_j)$, $\forall \mathbf{x}_i \in P$, $\forall \mathbf{x}_j \in Q$.
    \State Find perdicted labels $h_{t}(\mathbf{x}_j)$ using the SVM trained in round $t$, $\forall \mathbf{x}_j \in Q$.

\EndFor
\State Calculate $H(\mathbf{x}_j) = sign(\sum_{t=1}^T \alpha_{t}h_{t}(\mathbf{x}_j))$ $\forall \mathbf{x}_j \in Q$.

\State \hrulefill
\end{algorithmic}
\vspace{-3mm}
\end{figure}

\begin{figure}[!ht]
\vspace{-3mm}
\begin{center}

\subfloat[Case 1]{\includegraphics[width=0.17\textwidth, height=26mm]{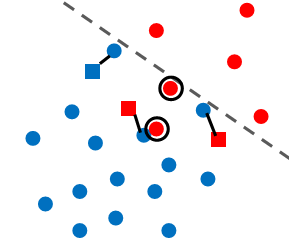}
\label{fig:ROI1}}
\quad \quad
\subfloat[Case 2]{\includegraphics[width=0.17\textwidth, height=26mm]{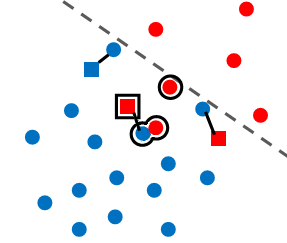}
\label{fig:ROI2}}

\subfloat[Case 3]{\includegraphics[width=0.17\textwidth, height=26mm]{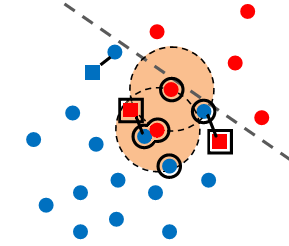}
\label{fig:ROI3}}
\quad
\subfloat[Figure Legends]{\includegraphics[width=0.21\textwidth, height=26mm]{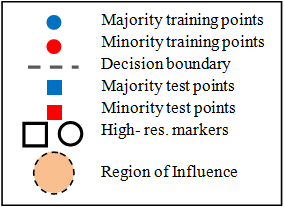}
\label{fig:ROIlegends}}

\caption{Effect of ROI on the estimation of resolution for test points.}
\label{fig:ROI}

\end{center}
\vspace{-4mm}
\end{figure}

Since KPBoost-SVM employs a datapoint-specific kernel perturbation scheme, the transformation factors for new points are unknown. Therefore, in KPBoost-SVM $D(\mathbf{x}_j)=1$ is used for all test points $\mathbf{x}_j$ in the test set $Q$. However, this causes the training and test points to reside in different spaces. A straight-forward solution is to assign to a test point, the transformation factor corresponding to its nearest training point. However, it is possible that points falling in the same vicinity of the input space but belonging to opposite classes are perturbed differently, resulting in erroneous estimation of transformation factors for some of the test points, in the absence of class information. This is illustrated in Figure \ref{fig:ROI1}. This difficulty can be tackled by ensuring that all points in the immediate vicinity of a misclassified point retain higher resolution in the next round. This would, however, only solve the problem partially as most test points, irrespective of their actual provenance, will still be assigned the transformation factors corresponding to points from the denser (usually majority) class. This situation is shown in Figure \ref{fig:ROI2} where only one of the two minority test points is assigned the correct resolution. Therefore, to further empower the sparser (usually minority) class, one must allow the misclassified points from the sparser class to impart high resolution to a wider Region Of Influence (ROI) compared to those of the denser class. It is clear from the illustration in Figure \ref{fig:ROI3} that assigning the transformation factor of the nearest training point makes sense when combined with the concept of ROI. Hence, here we slightly modify KPBoost-SVM to incorporate the concept of ROI, giving rise to the KPBoostROI-SVM technique detailed in Algorithm S1.

The ROI for the positive and negative classes, $ROI^{+}$ and $ROI^{-}$ respectively, are defined as
\begin{equation*}
\vspace{-1mm}
\begin{aligned}
& ROI^{+} = \vartheta \times \underset{\mathbf{x}_i \in \mathcal{C}_{+}^{P}}{\text{avg}} ( \underset{\underset{i' \neq i}{\mathbf{x}_{i'} \in \mathcal{C}_{+}^{P}}}{\min} (\|\mathbf{x}_i - \mathbf{x}_{i'}\|) ), \text{ and}\\
& ROI^{-} = \vartheta \times \underset{\mathbf{x}_i \in \mathcal{C}_{-}^{P}}{\text{avg}} ( \underset{\underset{i' \neq i}{\mathbf{x}_{i'} \in \mathcal{C}_{-}^{P}}}{\min} (\|\mathbf{x}_i - \mathbf{x}_{i'}\|) ),
\end{aligned}
\vspace{-2.5mm}
\end{equation*}
where $\mathcal{C}_{+}^{P}$ and $\mathcal{C}_{-}^{P}$ are respectively the sets of positive and negative training points while $\vartheta \in (0,1]$ is a user-defined ROI scaling parameter. In other words, the ROI for a particular class is chosen to be a fraction of the average nearest neighbor distance in the concerned class. Since the ROI values are based on the average nearest neighbor distances within the classes, the ROI for the sparser class will be larger than that of the denser class. Like in KPBoost-SVM, the set $\Omega$ corresponding to correctly classified points is identified after each round. Subsequently, the sets $\mho^{+}$ and $\mho^{-}$ of denominations of misclassified points respectively from the positive and negative classes are identified. Thereafter, all points within a $ROI^{+}$ radius of the points denominated in $\mho^{+}$ (denoted by the set $\Gamma^{+}$) or within a $ROI^{-}$ radius of those denominated in $\mho^{-}$ (denoted by $\Gamma^{-}$) are removed from $\Omega$, so as to retain high resolution for these points in the subsequent round. In other words, all correctly classified points within a $ROI^{+}$ (or $ROI^{-}$) radius of the misclassified positive (or negative) points are forced to retain high resolution. The rest of the training procedure is similar to that of KPBoost-SVM.

\subsubsection{Estimation of transformation factor and classification of a test point}
In order to classify a test point $\mathbf{x}_j$, the denomination $\Gamma_j$ pertaining to its nearest training point is identified. Then the transformation factor $D_t(\mathbf{x}_j)$ is calculated using the perturbation parameter $k_{\Gamma_j}^{t}$ corresponding to $\mathbf{x}_{\Gamma_j}$. The rest of the testing procedure is akin to that of KPBoost-SVM. An experimental comparison between KPBoost-SVM and KPBoostROI-SVM is reported in Section \ref{kpComp} of this supplementary document.

\section{Description of datasets}
A brief description highlighting the key properties of the datasets used in our experiments is presented in Table \ref{DataProp} (the alias used for each dataset in the subsequent tables is also reported). We have used a total of 52 two-class (42 low dimensional while the rest are high dimensional i.e. having more than 500 dimensions), and 21 multi-class (12 low dimensional and 9 high dimensional) datasets in this study. All the datasets are normalized so that each feature has zero mean and unit standard deviation. The chosen datasets (other than those from ImageNet) retain their original names with the suffixed numerals (if any) denoting either the target class only (in which case all the rest of the classes are combined together to form the non-target class) or the target as well as non-target classes \cite{triguero2017}. Apart from this some datasets required special construction and/or processing as listed in the following.
\begin{enumerate}
\item mnist\_2vs17 is constructed by randomly picking 100 images from the Special Dataset 1 and 500 images from the Special Dataset 3 \cite{lecun1998gradient} for each of the three classes 1, 2 and 7. mnist\_6vs09 is also constructed in a similar manner. 
\item rcv1\_1vs36vs245 is generated from Reuters RCV1 by randomly choosing 50 points from each class and using those of C15 as the first class, combining those of E21 and M11 to form the second class, and the rest as the third class.
\item We prepare 4 two-class and 6 multi-class dataset from ImageNet (2011 fall release) natural image classification database \cite{imagenet}. We download images from 8 primary subtrees or classes (plants, geological forms, natural objects, sports, artifacts, fungus, person and animal) and four leaves under the Miscellaneous branch (foods, microbes, collections and documents) ensuring every category to contain instances amounting at least 20 and at most 2\% of the number of synsets in the corresponding subtree. Instead of using raw images, a state-of-the-art feature representation is employed to express an image in the form of a 2048-dimensional real valued vector. We use the output of the final  global  average  pooling layer of Inception-v3 \cite{szegedy2016} deep learning network for the purpose. The two class datasets are created by combining images from two chosen ImageNet classes and named accordingly. imageNet3A (animal, artifacts and foods), imageNet3b (plant, artifacts and documents), imageNet4 (plants, artifacts, foods and documents), and imageNet8 (8 principal subtrees) are constructed by uniting instances from select classes. imageNet9 and imageNet12 are formed by incorporating all the downloaded images. However, in imageNet9, images from Miscellaneous branch are not distinctly labeled by the corresponding leaf name contrary to imageNet12.
\item breastcancer2 contains microRNA profiling of tissue samples collected immediately following surgery and 30 minutes after surgery from 14 patients only half of whom went through a radiotherapy treatment. A detailed description of the collection and processing of this dataset can be found in \cite{Fabris2016}.  
\end{enumerate}

\begin{table*}[!th]
\begin{center}
\scriptsize
\caption{ Summary of dataset properties. }
\label{DataProp}
\begin{tabular}{ c c c c c c c }
		\hline
		& & Number & Number & Number & Imbalance & Number \\
		Dataset & Alias & of & of & of & Ratio & of \\
		& & points & dimensions & classes & & disjuncts \\
		\hline
		abalone19  & TMD1 & 4177 & 8 & 2 & 129.53 & 233 \\
		abalone3vs11  & TMD2 & 502 & 8 & 2 & 32.46 & 34 \\
		abalone9vs18  & TMD3 & 731 & 8 & 2 & 16.40 & 69 \\
		banana  & TMD4 & 1213 & 2 & 2 & 4.09 & 51 \\
		car3  & TMD5 & 1728 & 6 & 2 & 24.04 & 83 \\
		cleveland0vs4  & TMD6 & 177 & 13 & 2 & 12.61 & 19 \\
		ecoli0137vs26  & TMD7 & 281 & 7 & 2 & 39.14 & 24 \\
		ecoli0vs1  & TMD8 & 220 & 7 & 2 & 1.85 & 20 \\
		ecoli1  & TMD9 & 336 & 7 & 2 & 3.36 & 28 \\
		ecoli2  & TMD10 & 336 & 7 & 2 & 5.46 & 20 \\
		ecoli3  & TMD11 & 336 & 7 & 2 & 8.60 & 23 \\
		ecoli4  & TMD12 & 336 & 7 & 2 & 15.80 & 23 \\
		glass015vs2  & TMD13 & 172 & 9 & 2 & 9.11 & 26 \\
		glass016vs2  & TMD14 & 192 & 9 & 2 & 10.29 & 35 \\
		glass04vs5  & TMD15 & 92 & 9 & 2 & 9.22 & 12 \\
		glass06vs5  & TMD16 & 108 & 9 & 2 & 11.00 & 24 \\
		glass0  & TMD17 & 213 & 9 & 2 & 2.04 & 35 \\
		glass123vs456  & TMD18 & 214 & 9 & 2 & 3.19 & 39 \\
		glass1  & TMD19 & 214 & 9 & 2 & 1.81 & 42 \\
		glass4  & TMD20 & 214 & 9 & 2 & 15.46 & 39 \\
		glass5  & TMD21 & 214 & 9 & 2 & 22.77 & 35 \\
		glass6\_aka\_glass7  & TMD22 & 214 & 9 & 2 & 6.37 & 33 \\
		iris12vs3  & TMD23 & 150 & 4 & 2 & 2.00 & 16 \\
		lymphography\_nf  & TMD24 & 148 & 18 & 2 & 23.66 & 17 \\
		pageblocks13vs4  & TMD25 & 472 & 10 & 2 & 15.85 & 19 \\
		pima  & TMD26 & 768 & 8 & 2 & 1.86 & 90 \\
		poker89vs6  & TMD27 & 1485 & 10 & 2 & 58.40 & 34 \\
		poker8vs6  & TMD28 & 1477 & 10 & 2 & 85.88 & 31 \\
		shuttle6vs23  & TMD29 & 230 & 9 & 2 & 22.00 & 29 \\
		shuttleC2vsC4  & TMD30 & 129 & 9 & 2 & 20.50 & 15 \\
		soybean12  & TMD31 & 683 & 35 & 2 & 14.52 & 64 \\
		thyroid1  & TMD32 & 215 & 5 & 2 & 5.14 & 20 \\
		vehicle0  & TMD33 & 846 & 18 & 2 & 3.25 & 33 \\
		vowel0  & TMD34 & 988 & 13 & 2 & 9.97 & 23 \\
		winequality\_red3vs5  & TMD35 & 691 & 11 & 2 & 68.10 & 67 \\
		winequality\_white3vs7  & TMD36 & 900 & 11 & 2 & 44.00 & 63 \\
		winequality\_white9vs4  & TMD37 & 168 & 11 & 2 & 32.60 & 26 \\
		yeast0359vs78  & TMD38 & 506 & 8 & 2 & 9.12 & 62 \\
		yeast0569vs4  & TMD39 & 528 & 8 & 2 & 9.35 & 49 \\
		yeast1458vs7  & TMD40 & 693 & 8 & 2 & 22.10 & 61 \\
		yeast2vs4  & TMD41 & 514 & 8 & 2 & 9.07 & 51 \\
		zoo3  & TMD42 & 101 & 16 & 2 & 19.20 & 17 \\
		\hline
		CNAE9\_2  & THD1 & 1080 & 856 & 2 & 8.00 & 430 \\
		CNAE9\_35vs6789  & THD2 & 360 & 856 & 2 & 2.00 & 187 \\
		CNAE9\_3vs4567  & THD3 & 480 & 856 & 2 & 3.00 & 210 \\
		breastcancer2  & THD4 & 28 & 2227 & 2 & 3.00 & 11 \\
		imageNet\_docs\_artifacts  & THD5 & 440 & 2048 & 2 & 21.00 & 77 \\
		imageNet\_food\_docs  & THD6 & 80 & 2048 & 2 & 3.00 & 12 \\
		imageNet\_plants\_artifacts  & THD7 & 600 & 2048 & 2 & 2.33 & 122 \\
		imageNet\_plants\_docs  & THD8 & 200 & 2048 & 2 & 9.00 & 43 \\
		mnist\_2vs17  & THD9 & 1800 & 784 & 2 & 2.00 & 183 \\
		mnist\_6vs09  & THD10 & 1800 & 784 & 2 & 2.00 & 148 \\
		\hline
		balance  & MMD1 & 625 & 4 & 3 & 5.87 & 118 \\
		dermatology  & MMD2 & 366 & 34 & 6 & 5.60 & 43 \\
		ecoli  & MMD3 & 336 & 7 & 8 & 71.50 & 136 \\
		glass2  & MMD4 & 214 & 10 & 6 & 8.44 & 29 \\
		hayes  & MMD5 & 132 & 4 & 3 & 1.70 & 36 \\
		lymphography  & MMD6 & 148 & 18 & 4 & 40.50 & 64 \\
		new-thyroid  & MMD7 & 215 & 5 & 3 & 5.00 & 24 \\
		pageblocks  & MMD8 & 548 & 10 & 5 & 164.00 & 81 \\
		shuttle  & MMD9 & 2175 & 9 & 5 & 853.00 & 657 \\
		thyroid  & MMD10 & 720 & 21 & 3 & 39.17 & 146 \\
		wine  & MMD11 & 178 & 13 & 3 & 1.47 & 23 \\
		yeast  & MMD12 & 1484 & 8 & 10 & 92.60 & 450 \\
		\hline
		CNAE9\_15vs249vs3vs6vs78  & MHD1 & 1080 & 856 & 5 & 3.00 & 462 \\
		CNAE9\_1vs2vs34vs56vs789  & MHD2 & 1080 & 856 & 5 & 3.00 & 452 \\
		imageNet12  & MHD3 & 1240 & 2048 & 12 & 21.00 & 214 \\
		imageNet8  & MHD4 & 1120 & 2048 & 8 & 21.00 & 170 \\
		imageNet9  & MHD5 & 1240 & 2048 & 9 & 21.00 & 210 \\
		imageNet3A  & MHD6 & 640 & 2048 & 3 & 7.00 & 110 \\
		imageNet3B  & MHD7 & 620 & 2048 & 3 & 21.00 & 95 \\
		imageNet4 & MHD8 & 680 & 2048 & 4 & 21.00 & 105 \\
		rcv1\_1vs36vs245  & MHD9 & 300 & 21513 & 3 & 3.00 & 272 \\
		\hline
\end{tabular}
\end{center}
\end{table*}

\begin{table*}[!th]
\begin{center}
\scriptsize
\caption{ Comparison of the performances of KPBoost-SVM and KPBoostROI-SVM on two-class datasets. }
\label{kpTwo}
\begin{tabular}{c | c c | c c | c c }
		\hline
		\multirow{2}{*}{\parbox{0.3cm}{\vspace{0.9cm} \rot{Datasets}}} & \multicolumn{2}{c|}{$Gmean$} & \multicolumn{2}{c|}{$AUC$} & \multicolumn{2}{c}{GSDI} \\ \cline{2-7}
		& \rot{KPBoost-SVM} & \rot{KPBoostROI-SVM} & \rot{KPBoost-SVM} & \rot{KPBoostROI-SVM} & \rot{KPBoost-SVM} & \rot{KPBoostROI-SVM} \\
		\hline
		TMD1 & \textbf{0.4159} & 0.2683 & \textbf{0.6000} & 0.5538 & \textbf{0.3712} & 0.2286 \\
		TMD2 & \textbf{1.0000} & \textbf{1.0000} & \textbf{1.0000} & \textbf{1.0000} & \textbf{1.0000} & \textbf{1.0000} \\
		TMD3 & \textbf{0.8133} & 0.8001 & \textbf{0.8234} & 0.8181 & \textbf{0.7195} & 0.6799 \\
		TMD4 & 0.9018 & \textbf{0.9033} & 0.9027 & \textbf{0.9059} & \textbf{0.4457} & 0.3149 \\
		TMD5 & \textbf{0.9846} & \textbf{0.9846} & \textbf{0.9849} & \textbf{0.9849} & \textbf{0.9940} & \textbf{0.9940} \\
		TMD6 & 0.8659 & \textbf{0.9022} & 0.8803 & \textbf{0.9136} & 0.5159 & \textbf{0.6817} \\
		TMD7 & 0.8590 & \textbf{0.8828} & 0.8712 & \textbf{0.9000} & \textbf{0.6885} & 0.6536 \\
		TMD8 & \textbf{0.9766} & \textbf{0.9766} & \textbf{0.9770} & \textbf{0.9770} & \textbf{0.8106} & \textbf{0.8106} \\
		TMD9 & \textbf{0.8919} & 0.8660 & \textbf{0.8930} & 0.8738 & 0.2669 & \textbf{0.2796} \\
		TMD10 & \textbf{0.9238} & 0.9160 & \textbf{0.9268} & 0.9194 & \textbf{0.4585} & \textbf{0.4585} \\
		TMD11 & \textbf{0.8527} & 0.8155 & \textbf{0.8675} & 0.8373 & \textbf{0.4983} & 0.3881 \\
		TMD12 & \textbf{0.9593} & 0.9182 & \textbf{0.9608} & 0.9234 & \textbf{0.7738} & 0.7147 \\
		TMD13 & \textbf{0.8527} & 0.7423 & \textbf{0.8559} & 0.7715 & \textbf{0.8054} & 0.6688 \\
		TMD14 & \textbf{0.7262} & 0.6323 & \textbf{0.7598} & 0.6876 & \textbf{0.5044} & 0.4875 \\
		TMD15 & \textbf{1.0000} & \textbf{1.0000} & \textbf{1.0000} & \textbf{1.0000} & \textbf{1.0000} & \textbf{1.0000} \\
		TMD16 & \textbf{1.0000} & \textbf{1.0000} & \textbf{1.0000} & \textbf{1.0000} & \textbf{1.0000} & \textbf{1.0000} \\
		TMD17 & \textbf{0.8135} & \textbf{0.8135} & \textbf{0.8192} & \textbf{0.8192} & \textbf{0.2095} & \textbf{0.2095} \\
		TMD18 & \textbf{0.9275} & 0.9237 & \textbf{0.9304} & 0.9257 & \textbf{0.7593} & 0.7227 \\
		TMD19 & \textbf{0.7224} & \textbf{0.7224} & \textbf{0.7431} & \textbf{0.7431} & \textbf{0.3933} & \textbf{0.3933} \\
		TMD20 & \textbf{0.8680} & \textbf{0.8680} & \textbf{0.8833} & \textbf{0.8833} & \textbf{0.7035} & \textbf{0.7035} \\
		TMD21 & 0.9951 & \textbf{0.9975} & 0.9951 & \textbf{0.9976} & 0.9854 & \textbf{0.9854} \\
		TMD22 & \textbf{0.9375} & \textbf{0.9375} & \textbf{0.9433} & \textbf{0.9433} & \textbf{0.5882} & \textbf{0.5882} \\
		TMD23 & \textbf{0.9847} & \textbf{0.9847} & \textbf{0.9850} & \textbf{0.9850} & \textbf{0.8310} & \textbf{0.8310} \\
		TMD24 & \textbf{1.0000} & \textbf{1.0000} & \textbf{1.0000} & \textbf{1.0000} & \textbf{1.0000} & \textbf{1.0000} \\
		TMD25 & \textbf{0.9977} & \textbf{0.9977} & \textbf{0.9978} & \textbf{0.9978} & 0.9028 & \textbf{0.9602} \\
		TMD26 & 0.7384 & \textbf{0.7443} & 0.7408 & \textbf{0.7486} & 0.4330 & \textbf{0.4638} \\
		TMD27 & \textbf{0.9935} & 0.9152 & \textbf{0.9935} & 0.9197 & \textbf{0.9922} & 0.9141 \\
		TMD28 & \textbf{1.0000} & \textbf{1.0000} & \textbf{1.0000} & \textbf{1.0000} & \textbf{1.0000} & \textbf{1.0000} \\
		TMD29 & \textbf{1.0000} & \textbf{1.0000} & \textbf{1.0000} & \textbf{1.0000} & \textbf{1.0000} & \textbf{1.0000} \\
		TMD30 & \textbf{1.0000} & \textbf{1.0000} & \textbf{1.0000} & \textbf{1.0000} & \textbf{1.0000} & \textbf{1.0000} \\
		TMD31 & \textbf{0.9771} & \textbf{0.9771} & \textbf{0.9778} & \textbf{0.9778} & \textbf{0.9681} & \textbf{0.9681} \\
		TMD32 & \textbf{0.9852} & 0.9703 & \textbf{0.9857} & 0.9714 & \textbf{1.0000} & 1.0000 \\
		TMD33 & \textbf{0.9847} & 0.9830 & \textbf{0.9848} & 0.9830 & 0.8766 & \textbf{0.9314} \\
		TMD34 & \textbf{1.0000} & \textbf{1.0000} & \textbf{1.0000} & \textbf{1.0000} & \textbf{1.0000} & \textbf{1.0000} \\
		TMD35 & \textbf{0.7101} & 0.2808 & \textbf{0.7496} & 0.5949 & \textbf{0.7642} & 0.2736 \\
		TMD36 & \textbf{0.7299} & 0.7239 & \textbf{0.7761} & 0.7682 & \textbf{0.7237} & 0.6997 \\
		TMD37 & \textbf{0.7969} & \textbf{0.7969} & \textbf{0.8970} & \textbf{0.8970} & \textbf{0.8000} & \textbf{0.8000} \\
		TMD38 & 0.5854 & \textbf{0.6354} & 0.6213 & \textbf{0.6850} & \textbf{0.5296} & 0.2051 \\
		TMD39 & \textbf{0.7883} & 0.7451 & \textbf{0.8010} & 0.7740 & \textbf{0.5479} & 0.4638 \\
		TMD40 & \textbf{0.5614} & 0.3593 & \textbf{0.6565} & 0.5985 & \textbf{0.5559} & 0.3603 \\
		TMD41 & \textbf{0.8776} & 0.8583 & \textbf{0.8847} & 0.8664 & \textbf{0.7246} & 0.4987 \\
		TMD42 & \textbf{0.9893} & 0.8000 & \textbf{0.9895} & 0.9000 & \textbf{0.9738} & 0.7738 \\
		\hline 
        \emph{Avg. Rank} & \textbf{1.3571} & 1.6429 & \textbf{1.3571} & 1.6429 & \textbf{1.3571} & 1.6429 \\
        \hline \hline
		THD1 & \textbf{0.9863} & 0.9781 & \textbf{0.9865} & 0.9786 & \textbf{0.9740} & 0.9604 \\
		THD2 & \textbf{1.0000} & \textbf{1.0000} & \textbf{1.0000} & \textbf{1.0000} & \textbf{1.0000} & \textbf{1.0000} \\
		THD3 & \textbf{1.0000} & \textbf{1.0000} & \textbf{1.0000} & \textbf{1.0000} & \textbf{0.9961} & \textbf{0.9961} \\
		THD4 & \textbf{0.6182} & \textbf{0.6182} & \textbf{0.7250} & \textbf{0.7250} & 0.0901 & \textbf{0.2231} \\
		THD5 & \textbf{1.0000} & 0.9414 & \textbf{1.0000} & 0.9500 & \textbf{1.0000} & 0.8436 \\
		THD6 & \textbf{1.0000} & \textbf{1.0000} & \textbf{1.0000} & \textbf{1.0000} & \textbf{1.0000} & \textbf{1.0000} \\
		THD7 & 0.9607 & \textbf{0.9715} & 0.9619 & \textbf{0.9722} & \textbf{0.8853} & 0.8763 \\
		THD8 & \textbf{1.0000} & \textbf{1.0000} & \textbf{1.0000} & \textbf{1.0000} & \textbf{1.0000} & \textbf{1.0000} \\
		THD9 & \textbf{0.9703} & \textbf{0.9703} & \textbf{0.9704} & \textbf{0.9704} & \textbf{0.8920} & \textbf{0.8920} \\
		THD10 & \textbf{0.9743} & \textbf{0.9743} & \textbf{0.9746} & \textbf{0.9746} & \textbf{0.7369} & \textbf{0.7369} \\ 
		\hline 
        \emph{Avg. Rank} & \textbf{1.4500} & 1.5500 & \textbf{1.4500} & 1.5500 & \textbf{1.4000} & 1.6000 \\
        \hline
\end{tabular}
\end{center}
\end{table*}

\section{Comparison of KPBoost-SVM and KPBoostROI-SVM}
\label{kpComp}
The comparison of the performances in terms of $Gmean$, $AUC$ and $GSDI$ of KPBoost-SVM and KPBoostROI-SVM on two-class and multi-class datasets are described in Table \ref{kpTwo} and Table \ref{kpMulti}, respectively. The reader should note that the parameter $\vartheta$ was varied in the set $\{0.6,0.7,0.8\}$ and the value of 0.6 was found to be the most effective. A closer inspection of Table \ref{kpTwo} reveals KPBoost-SVM to be a better imbalance resilient classifier than KPBoostROI-SVM on two-class datasets irrespective of dimensionality. From Table \ref{kpMulti} KPBoost-SVM using OVO decomposition is found to achieve better rank than KPBoostROI-SVM. However, in case of OVO, KPBoostROI-SVM can be considered as a better choice only on low dimensional datasets. Therefore, we deduce KPBoost-SVM to be the better method due to its consistently better scalable performance and algorithmic simplicity.   

\begin{table*}
\begin{center}
\scriptsize
\caption{ Comparison of the performances of KPBoost-SVM and KPBoostROI-SVM on multi-class datasets. }
\label{kpMulti}
	\begin{tabular}{ c | c c | c c | c c | c c | c c | c c }
		\hline
        \multirow{3}{*}{\parbox{0.3cm}{\vspace{0.9cm} \rot{Datasets}}} & \multicolumn{6}{c}{OVO} & \multicolumn{6}{|c}{OVA} \\ \cline{2-13}
		& \multicolumn{2}{c|}{$Gmean$} & \multicolumn{2}{c|}{$AUC$} & \multicolumn{2}{c|}{$GSDI$} & \multicolumn{2}{c|}{$Gmean$} & \multicolumn{2}{c|}{$AUC$} & \multicolumn{2}{c}{$GSDI$} \\ \cline{2-13}
		& \rot{KPBoost-SVM} & \rot{KPBoostROI-SVM} & \rot{KPBoost-SVM} & \rot{KPBoostROI-SVM} & \rot{KPBoost-SVM} & \rot{KPBoostROI-SVM} & \rot{KPBoost-SVM} & \rot{KPBoostROI-SVM} & \rot{KPBoost-SVM} & \rot{KPBoostROI-SVM} & \rot{KPBoost-SVM} & \rot{KPBoostROI-SVM} \\
		\hline
		MMD1 & \textbf{1.0000} & \textbf{1.0000} & \textbf{1.0000} & \textbf{1.0000} & \textbf{1.0000} & \textbf{1.0000} & 0.8527 & \textbf{0.8534} & 0.8931 & \textbf{0.8994} & 0.7806 & \textbf{0.8222} \\
		MMD2 & 0.9347 & \textbf{0.9548} & 0.9667 & \textbf{0.9747} & 0.0032 & \textbf{0.9464} & 0.9602 & \textbf{0.9629} & 0.9769 & \textbf{0.9783} & \textbf{0.8637} & 0.7950 \\
		MMD3 & \textbf{0.0000} & \textbf{0.0000} & \textbf{0.8243} & \textbf{0.8243} & \textbf{0.0000} & \textbf{0.0000} & \textbf{0.0000} & \textbf{0.0000} & \textbf{0.8946} & 0.8892 & \textbf{0.0000} & \textbf{0.0000} \\
		MMD4 & \textbf{0.9388} & \textbf{0.9388} & \textbf{0.9663} & \textbf{0.9663} & \textbf{0.8134} & \textbf{0.8134} & \textbf{0.8443} & 0.7458 & \textbf{0.9203} & 0.8909 & \textbf{0.3568} & 0.2832 \\
		MMD5 & \textbf{0.8892} & 0.7299 & \textbf{0.9219} & 0.8316 & \textbf{0.7801} & 0.4420 & \textbf{0.8486} & 0.8117 & \textbf{0.8941} & 0.8663 & \textbf{0.6748} & 0.6748 \\
		MMD6 & \textbf{0.9232} & \textbf{0.9232} & \textbf{0.9516} & \textbf{0.9516} & \textbf{0.8732} & \textbf{0.8732} & \textbf{0.9330} & \textbf{0.9330} & \textbf{0.9582} & \textbf{0.9582} & \textbf{0.8859} & \textbf{0.8859} \\
		MMD7 & \textbf{0.9353} & \textbf{0.9353} & \textbf{0.9545} & \textbf{0.9545} & \textbf{0.9353} & \textbf{0.9353} & 0.9687 & \textbf{1.0000} & 0.9773 & \textbf{1.0000} & 0.9687 & \textbf{1.0000} \\
		MMD8 & \textbf{0.8292} & \textbf{0.8292} & \textbf{0.9100} & \textbf{0.9100} & \textbf{0.7454} & \textbf{0.7454} & 0.8992 & \textbf{0.9524} & 0.9412 & \textbf{0.9725} & 0.5963 & \textbf{0.7754} \\
		MMD9 & \textbf{1.0000} & \textbf{1.0000} & \textbf{1.0000} & \textbf{1.0000} & \textbf{1.0000} & \textbf{1.0000} & 0.0000 & \textbf{0.8706} & 0.8091 & \textbf{0.9375} & 0.0000 & \textbf{0.8706} \\
		MMD10 & \textbf{0.9583} & 0.8285 & \textbf{0.9700} & 0.8896 & \textbf{0.9291} & 0.8007 & \textbf{0.8224} & 0.5466 & \textbf{0.8683} & 0.7033 & \textbf{0.8726} & 0.4443 \\
		MMD11 & \textbf{0.9773} & \textbf{0.9773} & \textbf{0.9833} & \textbf{0.9833} & \textbf{1.0000} & \textbf{1.0000} & \textbf{1.0000} & 0.9846 & \textbf{1.0000} & 0.9886 & \textbf{1.0000} & 0.9119 \\
		MMD12 & \textbf{0.0000} & \textbf{0.0000} & 0.7858 & \textbf{0.7863} & \textbf{0.0000} & \textbf{0.0000} & 0.3601 & \textbf{0.4467} & 0.6621 & \textbf{0.7289} & 0.0368 & \textbf{0.0425} \\
		\hline
		\emph{Avg Rank} & \textbf{1.46} & 1.54 & \textbf{1.50} & \textbf{1.50} & \textbf{1.46}	& 1.54 & 1.58 & \textbf{1.42} & 1.54 & \textbf{1.46} &	 \textbf{1.50} & \textbf{1.50} \\ \hline \hline
		MHD1 & 0.9288 & \textbf{0.9449} & 0.9574 & \textbf{0.9666} & 0.9148 & \textbf{0.9361} & 0.9282 & \textbf{0.9293} & 0.9556 & \textbf{0.9568} & \textbf{0.9186} & 0.9039 \\
		MHD2 & 0.9340 & \textbf{0.9352} & 0.9595 & \textbf{0.9601} & \textbf{0.8908} & 0.8790 & 0.9460 & \textbf{0.9503} & 0.9665 & \textbf{0.9694} & \textbf{0.8949} & 0.8833 \\
		MHD3 & \textbf{0.0000} & \textbf{0.0000} & \textbf{0.8042} & 0.8042 & \textbf{0.0000} & \textbf{0.0000} & \textbf{0.6405} & 0.0000 & \textbf{0.8321} & 0.8210 & \textbf{0.1794} & 0.0000 \\
		MHD4 & \textbf{0.0000} & \textbf{0.0000} & \textbf{0.8475} & 0.8254 & \textbf{0.0000} & \textbf{0.0000} & \textbf{0.7255} & 0.7028 & \textbf{0.8641} & 0.8492 & \textbf{0.3206} & 0.1788 \\
		MHD5 & \textbf{0.0000} & \textbf{0.0000} & \textbf{0.7897} & \textbf{0.7897} & \textbf{0.0000} & \textbf{0.0000} & \textbf{0.6971} & 0.6482 & \textbf{0.8755} & 0.8331 & 0.3016 & \textbf{0.3025} \\
		MHD6 & \textbf{0.8595} & \textbf{0.8595} & \textbf{0.9048} & \textbf{0.9048} & \textbf{0.6638} & \textbf{0.6638} & 0.8441 & \textbf{0.8838} & 0.8919 & \textbf{0.9157} & 0.6808 & \textbf{0.8396} \\
		MHD7 & \textbf{0.9880} & \textbf{0.9880} & \textbf{0.9911} & \textbf{0.9911} & \textbf{0.9527} & \textbf{0.9527} & \textbf{0.9880} & 0.9756 & \textbf{0.9911} & 0.9821 & \textbf{0.9527} & 0.9001 \\
		MHD8 & \textbf{0.8660} & 0.6777 & \textbf{0.9173} & 0.8247 & \textbf{0.8267} & 0.0040 & \textbf{0.9219} & 0.8613 & \textbf{0.9511} & 0.9140 & \textbf{0.0048} & 0.0042 \\
		MHD9 & \textbf{0.7511} & \textbf{0.7511} & \textbf{0.8361} & \textbf{0.8361} & \textbf{0.7076} & \textbf{0.7076} & \textbf{0.7556} & 0.7147 & \textbf{0.8333} & 0.8000 & \textbf{0.7231} & 0.6502 \\
		\hline
		\emph{Avg. Rank} & 1.56	 & \textbf{1.44} & \textbf{1.44} & 1.56	& \textbf{1.44} & 1.56 &	\textbf{1.33} &  1.67 & \textbf{1.33} & 1.67 &	\textbf{1.22} & 1.78 \\ \hline

\end{tabular}
\end{center}
\end{table*}

\section{Detailed results}
The detailed performance of the proposed KPBoost-SVM against the rest of the state-of-the-art contenders (namely SVM, AbaBoost-SVM, RUSBoost, C4.5/IB1, D-SVM, AKS, AKS-$\chi^2$, AdaBoost-MLP, KBA, and ACT) in terms of $Gmean$, $AUC$ and $GSDI$ on the two-class as well as multi-class datasets are detailed in Tables \ref{twoGM}-\ref{multiGSDI}.

\begin{table*}[!th]
\begin{center}
\scriptsize
\caption{ Results on two-class datasets in terms of $Gmean$. }
\label{twoGM}
\begin{tabular}{c | c | c | c | c | c | c | c | c | c | c | c }
		\hline
		\rot{Datasets} & \rot{KPBoost-SVM} & \rot{SVM} & \rot{AdaBoost-SVM} & \rot{C4.5/IB1} & \rot{D-SVM} & \rot{AKS} & \rot{AKS-$\chi^2$} & \rot{RUSBoost} & \rot{AdaBoost-MLP} & \rot{KBA} & \rot{ACT} \\
		\hline
		TMD1 & 0.4159 & 0.1565 & 0.0754 & 0.0000 & 0.0000 & \textbf{0.5594} & 0.1561 & 0.0731 & 0.0000 & 0.0754 & 0.5103 \\
		TMD2 & \textbf{1.0000} & \textbf{1.0000} & \textbf{1.0000} & 0.9633 & \textbf{1.0000} & \textbf{1.0000} & \textbf{1.0000} & 0.0000 & \textbf{1.0000} & \textbf{1.0000} & \textbf{1.0000} \\
		TMD3 & \textbf{0.8133} & 0.6040 & 0.4000 & 0.3345 & 0.4942 & 0.6336 & 0.4631 & 0.5086 & 0.6884 & 0.5329 & 0.3813 \\
		TMD4 & \textbf{0.9018} & 0.8056 & 0.0000 & 0.8207 & 0.2612 & 0.7478 & 0.7076 & 0.8696 & 0.8635 & 0.5671 & 0.1130 \\
		TMD5 & 0.9846 & 0.9688 & 0.8605 & 0.0000 & 0.9462 & \textbf{0.9982} & 0.9752 & 0.9497 & 0.9758 & 0.9289 & 0.9009 \\
		TMD6 & 0.8659 & 0.8634 & 0.0000 & 0.3446 & 0.8547 & 0.9001 & \textbf{0.9025} & 0.8551 & 0.8241 & 0.7813 & 0.8634 \\
		TMD7 & 0.8590 & \textbf{0.8828} & 0.1982 & 0.7359 & \textbf{0.8828} & \textbf{0.8828} & 0.8810 & 0.7346 & 0.8797 & \textbf{0.8828} & \textbf{0.8828} \\
		TMD8 & 0.9766 & 0.9461 & 0.9524 & \textbf{0.9831} & 0.9557 & 0.9539 & 0.9317 & 0.9761 & 0.9660 & 0.7504 & 0.9461 \\
		TMD9 & \textbf{0.8919} & 0.8428 & 0.3425 & 0.7881 & 0.8111 & 0.7750 & 0.4826 & 0.8624 & 0.8405 & 0.7269 & 0.0000 \\
		TMD10 & 0.9238 & 0.8952 & 0.1965 & 0.7472 & 0.8877 & \textbf{0.9321} & 0.7593 & 0.8760 & 0.8609 & 0.7401 & 0.0000 \\
		TMD11 & \textbf{0.8527} & 0.7322 & 0.0000 & 0.4999 & 0.4708 & 0.7838 & 0.2703 & 0.7282 & 0.7429 & 0.7614 & 0.4000 \\
		TMD12 & 0.9593 & 0.8025 & 0.0000 & 0.6533 & 0.7081 & \textbf{0.9790} & 0.7293 & 0.8859 & 0.8810 & 0.3857 & 0.7182 \\
		TMD13 & \textbf{0.8527} & 0.5366 & 0.0000 & 0.0000 & 0.0000 & 0.7966 & 0.3212 & 0.2277 & 0.3850 & 0.5629 & 0.2298 \\
		TMD14 & 0.7262 & 0.4151 & 0.0000 & 0.0000 & 0.0926 & \textbf{0.7787} & 0.4199 & 0.0000 & 0.2511 & 0.1604 & 0.0956 \\
		TMD15 & \textbf{1.0000} & \textbf{1.0000} & 0.1414 & 0.9940 & 0.9295 & \textbf{1.0000} & 0.9312 & 0.7095 & 0.7940 & 0.8842 & 0.3414 \\
		TMD16 & \textbf{1.0000} & \textbf{1.0000} & 0.1414 & 0.2793 & 0.9364 & \textbf{1.0000} & 0.9847 & 0.9364 & 0.9947 & 0.7414 & 0.4372 \\
		TMD17 & 0.8135 & 0.7708 & 0.0756 & 0.0000 & 0.6509 & 0.7969 & 0.6065 & 0.3430 & 0.7247 & 0.6819 & \textbf{0.8418} \\
		TMD18 & \textbf{0.9275} & 0.8170 & 0.8072 & 0.6452 & 0.7454 & 0.6728 & 0.7203 & 0.8212 & 0.9072 & 0.7304 & 0.3435 \\
		TMD19 & \textbf{0.7224} & 0.7002 & 0.0000 & 0.0516 & 0.7068 & 0.7093 & 0.7120 & 0.0962 & 0.6871 & 0.6380 & 0.6357 \\
		TMD20 & 0.8680 & 0.6899 & 0.2569 & 0.6715 & 0.6899 & 0.8680 & \textbf{0.9329} & 0.4952 & 0.8589 & 0.5942 & 0.8956 \\
		TMD21 & \textbf{0.9951} & 0.9414 & 0.0000 & 0.0000 & 0.9341 & \textbf{0.9951} & 0.2811 & 0.5276 & 0.7951 & 0.7975 & 0.3975 \\
		TMD22 & \textbf{0.9375} & 0.9008 & 0.9295 & 0.8880 & 0.9151 & 0.9320 & 0.9308 & 0.9126 & 0.9096 & 0.8782 & 0.9241 \\
		TMD23 & \textbf{0.9847} & 0.9536 & 0.9795 & 0.8606 & 0.9747 & 0.9530 & 0.6148 & 0.7636 & 0.9317 & 0.6309 & 0.7596 \\
		TMD24 & \textbf{1.0000} & 0.8000 & 0.7414 & 0.5341 & 0.6000 & \textbf{1.0000} & 0.9749 & 0.5281 & 0.5965 & 0.5414 & 0.8000 \\
		TMD25 & 0.9977 & 0.8960 & 0.1265 & 0.8986 & 0.8619 & 0.9581 & 0.9735 & 0.7966 & \textbf{0.9989} & 0.7494 & 0.9814 \\
		TMD26 & \textbf{0.7384} & 0.7137 & 0.6940 & 0.5035 & 0.6306 & 0.5896 & 0.7100 & 0.5686 & 0.7359 & 0.5606 & 0.4255 \\
		TMD27 & \textbf{0.9935} & 0.8631 & 0.0000 & 0.0000 & 0.6893 & 0.9312 & 0.7941 & 0.0000 & 0.8941 & 0.7374 & 0.2159 \\
		TMD28 & \textbf{1.0000} & \textbf{1.0000} & 0.2000 & 0.0000 & 0.4023 & 0.9997 & 0.7249 & 0.0000 & 0.8992 & 0.6680 & 0.0000 \\
		TMD29 & \textbf{1.0000} & \textbf{1.0000} & 0.4828 & 0.9954 & 0.9414 & \textbf{1.0000} & 0.9375 & 0.7908 & 0.9414 & 0.8828 & \textbf{1.0000} \\
		TMD30 & \textbf{1.0000} & 0.8000 & 0.1960 & 0.6000 & 0.7960 & \textbf{1.0000} & 0.9874 & 0.0000 & 0.8000 & 0.8000 & \textbf{1.0000} \\
		TMD31 & 0.9771 & 0.9771 & 0.8926 & 0.9603 & 0.9649 & 0.9945 & 0.9771 & \textbf{0.9992} & 0.9519 & 0.9771 & 0.9771 \\
		TMD32 & 0.9852 & 0.9555 & 0.7892 & 0.8272 & 0.9771 & 0.9888 & 0.2600 & \textbf{0.9916} & 0.9499 & 0.5703 & 0.5972 \\
		TMD33 & \textbf{0.9847} & 0.9751 & 0.5892 & 0.7023 & 0.8787 & 0.9841 & 0.4538 & 0.9181 & 0.9761 & 0.5459 & 0.3415 \\
		TMD34 & \textbf{1.0000} & \textbf{1.0000} & 0.8648 & 0.8006 & \textbf{1.0000} & \textbf{1.0000} & \textbf{1.0000} & 0.9330 & \textbf{1.0000} & \textbf{1.0000} & \textbf{1.0000} \\
		TMD35 & \textbf{0.7101} & 0.1414 & 0.0000 & 0.0000 & 0.1109 & 0.4848 & 0.1404 & 0.6050 & 0.1404 & 0.0874 & 0.0000 \\
		TMD36 & 0.7299 & 0.6461 & 0.0997 & 0.4406 & 0.5306 & \textbf{0.7446} & 0.5112 & 0.6860 & 0.5237 & 0.5443 & 0.0000 \\
		TMD37 & 0.7969 & \textbf{0.7969} & 0.0000 & 0.1936 & \textbf{0.7969} & \textbf{0.7969} & 0.5908 & 0.7490 & 0.5969 & 0.4000 & 0.7908 \\
		TMD38 & \textbf{0.5854} & 0.4068 & 0.4068 & 0.3035 & 0.2553 & 0.4680 & 0.3242 & 0.1230 & 0.4172 & 0.3074 & 0.0000 \\
		TMD39 & \textbf{0.7883} & 0.5550 & 0.4582 & 0.2750 & 0.0000 & 0.4918 & 0.5492 & 0.3945 & 0.6124 & 0.6013 & 0.0000 \\
		TMD40 & 0.5614 & 0.1971 & 0.0000 & 0.0000 & 0.0000 & \textbf{0.6587} & 0.1617 & 0.0000 & 0.0000 & 0.0807 & 0.0000 \\
		TMD41 & \textbf{0.8776} & 0.8121 & 0.1524 & 0.7677 & 0.5601 & 0.4838 & 0.6533 & 0.8594 & 0.8610 & 0.6337 & 0.6209 \\
		TMD42 & \textbf{0.9893} & 0.6000 & 0.4000 & 0.0000 & 0.5663 & 0.8000 & 0.5947 & 0.3947 & 0.5947 & 0.6000 & 0.5947 \\
		\hline
		THD1 & \textbf{0.9863} & 0.9565 & 0.0000 & 0.9654 & 0.9594 & 0.9852 & 0.8952 & 0.9482 & 0.9701 & 0.8192 & 0.0000 \\
		THD2 & \textbf{1.0000} & 0.9874 & 0.0000 & 0.9550 & 0.9787 & 0.8403 & 0.9249 & 0.9595 & 0.9789 & 0.9874 & 0.1239 \\
		THD3 & \textbf{1.0000} & 0.9873 & 0.0000 & 0.9115 & 0.9668 & 0.9731 & 0.8802 & 0.9775 & 0.9874 & 0.9873 & 0.0816 \\
		THD4 & \textbf{0.6182} & 0.1732 & 0.0000 & 0.1732 & 0.0000 & 0.5602 & 0.5016 & 0.0000 & 0.2957 & 0.0000 & 0.3864 \\
		THD5 & \textbf{1.0000} & 0.6243 & 0.6243 & 0.9325 & 0.6243 & \textbf{1.0000} & 0.0000 & 0.7535 & \textbf{1.0000} & 0.1414 & 0.0000 \\
		THD6 & \textbf{1.0000} & 0.9414 & 0.9414 & 0.6788 & 0.9414 & \textbf{1.0000} & 0.0000 & 0.4446 & 0.8000 & 0.1414 & 0.0000 \\
		THD7 & 0.9607 & 0.9362 & 0.8992 & 0.8248 & 0.9575 & \textbf{0.9709} & 0.5360 & 0.8804 & 0.9209 & 0.9112 & 0.9060 \\
		THD8 & \textbf{1.0000} & 0.7414 & 0.7414 & 0.4665 & 0.7414 & \textbf{1.0000} & 0.0000 & 0.6658 & \textbf{1.0000} & 0.1414 & 0.0000 \\
		THD9 & 0.9703 & 0.9669 & 0.0000 & 0.1314 & 0.9631 & 0.9749 & 0.9573 & 0.1314 & \textbf{0.9816} & 0.9677 & 0.9310 \\
		THD10 & 0.9743 & 0.9726 & 0.0000 & 0.1311 & 0.9825 & 0.9837 & 0.9715 & 0.1311 & 0.9921 & 0.9657 & \textbf{0.9946} \\
		\hline
\end{tabular}
\end{center}
\end{table*}

\begin{table*}[!th]
\begin{center}
\scriptsize
\caption{ Results of two-class datasets in terms of $AUC$. }
\begin{tabular}{c | c | c | c | c | c | c | c | c | c | c | c } \hline
		\rot{Datasets} & \rot{KPBoost-SVM} & \rot{SVM} & \rot{AdaBoost-SVM} & \rot{C4.5/IB1} & \rot{D-SVM} & \rot{AKS} & \rot{AKS-$\chi^2$} & \rot{RUSBoost} & \rot{AdaBoost-MLP} & \rot{KBA} & \rot{ACT} \\ \hline
		TMD1 & 0.6000 & 0.5266 & 0.5122 & 0.5000 & 0.4967 & \textbf{0.6435} & 0.5235 & 0.4961 & 0.5000 & 0.5118 & 0.6108 \\
		TMD2 & \textbf{1.0000} & \textbf{1.0000} & \textbf{1.0000} & 0.9667 & \textbf{1.0000} & \textbf{1.0000} & \textbf{1.0000} & 0.5000 & \textbf{1.0000} & \textbf{1.0000} & \textbf{1.0000} \\
		TMD3 & \textbf{0.8234} & 0.6908 & 0.6005 & 0.5693 & 0.6574 & 0.7132 & 0.6415 & 0.6133 & 0.7376 & 0.6226 & 0.6228 \\
		TMD4 & \textbf{0.9027} & 0.8230 & 0.5000 & 0.8298 & 0.5022 & 0.7892 & 0.7495 & 0.8709 & 0.8690 & 0.5872 & 0.5319 \\
		TMD5 & 0.9849 & 0.9697 & 0.8716 & 0.5000 & 0.9477 & \textbf{0.9982} & 0.9759 & 0.9501 & 0.9762 & 0.9329 & 0.9038 \\
		TMD6 & 0.8803 & 0.8773 & 0.5000 & 0.5879 & 0.8682 & \textbf{0.9106} & 0.9091 & 0.8649 & 0.8409 & 0.8136 & 0.8773 \\
		TMD7 & 0.8712 & \textbf{0.9000} & 0.5982 & 0.8445 & \textbf{0.9000} & \textbf{0.9000} & 0.8982 & 0.8427 & 0.8963 & \textbf{0.9000} & \textbf{0.9000} \\
		TMD8 & 0.9770 & 0.9479 & 0.9542 & \textbf{0.9832} & 0.9559 & 0.9551 & 0.9346 & 0.9763 & 0.9665 & 0.8133 & 0.9479 \\
		TMD9 & \textbf{0.8930} & 0.8563 & 0.6166 & 0.8047 & 0.8173 & 0.8081 & 0.6483 & 0.8663 & 0.8472 & 0.7561 & 0.5000 \\
		TMD10 & 0.9268 & 0.9013 & 0.5965 & 0.7875 & 0.8930 & \textbf{0.9350} & 0.7898 & 0.8854 & 0.8760 & 0.7733 & 0.5000 \\
		TMD11 & \textbf{0.8675} & 0.7709 & 0.5000 & 0.6665 & 0.6818 & 0.8106 & 0.6060 & 0.7635 & 0.7751 & 0.7887 & 0.6330 \\
		TMD12 & 0.9608 & 0.8250 & 0.5000 & 0.7219 & 0.8124 & \textbf{0.9794} & 0.7750 & 0.8921 & 0.8921 & 0.6219 & 0.8234 \\
		TMD13 & \textbf{0.8559} & 0.6742 & 0.5000 & 0.5000 & 0.5000 & 0.8129 & 0.6172 & 0.5272 & 0.6030 & 0.7043 & 0.5731 \\
		TMD14 & 0.7598 & 0.6326 & 0.5000 & 0.5000 & 0.5107 & \textbf{0.7981} & 0.6412 & 0.5000 & 0.5605 & 0.5464 & 0.5019 \\
		TMD15 & \textbf{1.0000} & \textbf{1.0000} & 0.5500 & 0.9941 & 0.9382 & \textbf{1.0000} & 0.9382 & 0.8206 & 0.8941 & 0.8890 & 0.6500 \\
		TMD16 & \textbf{1.0000} & \textbf{1.0000} & 0.5500 & 0.5900 & 0.9450 & \textbf{1.0000} & 0.9850 & 0.9450 & 0.9947 & 0.8500 & 0.6416 \\
		TMD17 & 0.8192 & 0.7871 & 0.5143 & 0.5000 & 0.6962 & 0.8084 & 0.6376 & 0.5032 & 0.7309 & 0.7255 & \textbf{0.8451} \\
		TMD18 & \textbf{0.9304} & 0.8379 & 0.8288 & 0.7236 & 0.7861 & 0.7464 & 0.7727 & 0.8444 & 0.9117 & 0.7658 & 0.6438 \\
		TMD19 & \textbf{0.7431} & 0.7268 & 0.5000 & 0.5067 & 0.7299 & 0.7334 & 0.7200 & 0.5176 & 0.7037 & 0.6802 & 0.6853 \\
		TMD20 & 0.8833 & 0.8000 & 0.5833 & 0.7258 & 0.8000 & 0.8833 & \textbf{0.9352} & 0.7068 & 0.8734 & 0.7333 & 0.9067 \\
		TMD21 & \textbf{0.9951} & 0.9500 & 0.5000 & 0.4976 & 0.9427 & \textbf{0.9951} & 0.5976 & 0.7329 & 0.8951 & 0.8232 & 0.6976 \\
		TMD22 & \textbf{0.9433} & 0.9100 & 0.9352 & 0.8938 & 0.9213 & 0.9379 & 0.9357 & 0.9186 & 0.9159 & 0.8879 & 0.9273 \\
		TMD23 & \textbf{0.9850} & 0.9550 & 0.9800 & 0.8700 & 0.9750 & 0.9550 & 0.7300 & 0.8650 & 0.9350 & 0.6750 & 0.8600 \\
		TMD24 & \textbf{1.0000} & 0.9000 & 0.8500 & 0.7429 & 0.8000 & \textbf{1.0000} & 0.9754 & 0.7358 & 0.7966 & 0.7500 & 0.9000 \\
		TMD25 & 0.9978 & 0.9055 & 0.5400 & 0.9077 & 0.8811 & 0.9600 & 0.9744 & 0.8966 & \textbf{0.9989} & 0.7684 & 0.9822 \\
		TMD26 & 0.7408 & 0.7335 & 0.7164 & 0.6135 & 0.6764 & 0.6621 & 0.7227 & 0.6404 & \textbf{0.7412} & 0.6578 & 0.6365 \\
		TMD27 & \textbf{0.9935} & 0.8800 & 0.5000 & 0.5000 & 0.7400 & 0.9326 & 0.8200 & 0.4928 & 0.8997 & 0.7684 & 0.5600 \\
		TMD28 & \textbf{1.0000} & \textbf{1.0000} & 0.6000 & 0.5000 & 0.5062 & 0.9997 & 0.7667 & 0.4884 & 0.9076 & 0.7833 & 0.5000 \\
		TMD29 & \textbf{1.0000} & \textbf{1.0000} & 0.7000 & 0.9955 & 0.9500 & \textbf{1.0000} & 0.9455 & 0.8909 & 0.9500 & 0.9000 & \textbf{1.0000} \\
		TMD30 & \textbf{1.0000} & 0.9000 & 0.5960 & 0.8000 & 0.8960 & \textbf{1.0000} & 0.9877 & 0.5000 & 0.9000 & 0.9000 & \textbf{1.0000} \\
		TMD31 & 0.9778 & 0.9778 & 0.8986 & 0.9620 & 0.9667 & 0.9945 & 0.9778 & \textbf{0.9992} & 0.9556 & 0.9778 & 0.9778 \\
		TMD32 & 0.9857 & 0.9571 & 0.8143 & 0.8405 & 0.9774 & 0.9889 & 0.5829 & \textbf{0.9917} & 0.9516 & 0.6714 & 0.7972 \\
		TMD33 & \textbf{0.9848} & 0.9752 & 0.6778 & 0.7458 & 0.8831 & 0.9841 & 0.5706 & 0.9187 & 0.9762 & 0.6188 & 0.5449 \\
		TMD34 & \textbf{1.0000} & \textbf{1.0000} & 0.8744 & 0.8211 & \textbf{1.0000} & \textbf{1.0000} & \textbf{1.0000} & 0.9355 & \textbf{1.0000} & \textbf{1.0000} & \textbf{1.0000} \\
		TMD35 & \textbf{0.7496} & 0.5493 & 0.4993 & 0.5000 & 0.5147 & 0.6541 & 0.5463 & 0.7236 & 0.5463 & 0.5074 & 0.4000 \\
		TMD36 & 0.7761 & 0.7244 & 0.5244 & 0.6233 & 0.6062 & \textbf{0.7886} & 0.6693 & 0.7347 & 0.6739 & 0.6835 & 0.5000 \\
		TMD37 & \textbf{0.8970} & \textbf{0.8970} & 0.5000 & 0.5907 & 0.8939 & \textbf{0.8970} & 0.7848 & 0.8330 & 0.7939 & 0.7000 & 0.8879 \\
		TMD38 & 0.6213 & 0.6067 & 0.6067 & 0.5734 & 0.5243 & \textbf{0.6251} & 0.5878 & 0.5090 & 0.6134 & 0.5778 & 0.5000 \\
		TMD39 & \textbf{0.8010} & 0.6611 & 0.6089 & 0.5461 & 0.5000 & 0.6790 & 0.6602 & 0.5854 & 0.6911 & 0.6568 & 0.5000 \\
		TMD40 & 0.6565 & 0.5462 & 0.5000 & 0.5000 & 0.5000 & \textbf{0.6719} & 0.5311 & 0.4381 & 0.4970 & 0.5136 & 0.5000 \\
		TMD41 & \textbf{0.8847} & 0.8312 & 0.5568 & 0.8031 & 0.7015 & 0.6500 & 0.7344 & 0.8687 & 0.8686 & 0.6939 & 0.7412 \\
		TMD42 & \textbf{0.9895} & 0.8000 & 0.7000 & 0.5000 & 0.7684 & 0.9000 & 0.7947 & 0.6789 & 0.7947 & 0.8000 & 0.7947 \\
		\hline
		THD1 & \textbf{0.9865} & 0.9583 & 0.5000 & 0.9661 & 0.9620 & 0.9854 & 0.9021 & 0.9495 & 0.9708 & 0.8370 & 0.5000 \\
		THD2 & \textbf{1.0000} & 0.9875 & 0.5000 & 0.9563 & 0.9792 & 0.8542 & 0.9271 & 0.9604 & 0.9792 & 0.9875 & 0.5250 \\
		THD3 & \textbf{1.0000} & 0.9875 & 0.5000 & 0.9153 & 0.9681 & 0.9736 & 0.8875 & 0.9778 & 0.9875 & 0.9875 & 0.5083 \\
		THD4 & \textbf{0.7250} & 0.5500 & 0.5000 & 0.4250 & 0.5000 & 0.6850 & 0.5950 & 0.4000 & 0.5500 & 0.5000 & 0.5800 \\
		THD5 & \textbf{1.0000} & 0.7500 & 0.7500 & 0.9405 & 0.7500 & \textbf{1.0000} & 0.5000 & 0.7738 & \textbf{1.0000} & 0.5500 & 0.5000 \\
		THD6 & \textbf{1.0000} & 0.9500 & 0.9500 & 0.7833 & 0.9500 & \textbf{1.0000} & 0.5000 & 0.6333 & 0.9000 & 0.5500 & 0.5000 \\
		THD7 & 0.9619 & 0.9389 & 0.9056 & 0.8294 & 0.9587 & \textbf{0.9714} & 0.6444 & 0.8817 & 0.9238 & 0.9167 & 0.9151 \\
		THD8 & \textbf{1.0000} & 0.8500 & 0.8500 & 0.6667 & 0.8500 & \textbf{1.0000} & 0.5000 & 0.7833 & \textbf{1.0000} & 0.5500 & 0.5000 \\
		THD9 & 0.9704 & 0.9671 & 0.5000 & 0.5092 & 0.9633 & 0.9750 & 0.9575 & 0.5092 & \textbf{0.9817} & 0.9679 & 0.9333 \\
		THD10 & 0.9746 & 0.9729 & 0.5000 & 0.5100 & 0.9825 & 0.9838 & 0.9717 & 0.5100 & 0.9921 & 0.9662 & \textbf{0.9946} \\
		\hline
\end{tabular}
\end{center}
\end{table*}

\begin{table*}[!th]
\begin{center}
\scriptsize
\caption{ Results on two-class datasets in terms of $GSDI$. }
\begin{tabular}{c | c | c | c | c | c | c | c | c | c | c | c }
		\hline
		\rot{Datasets} & \rot{KPBoost-SVM} & \rot{SVM} & \rot{AdaBoost-SVM} & \rot{C4.5/IB1} & \rot{D-SVM} & \rot{AKS} & \rot{AKS-$\chi^2$} & \rot{RUSBoost} & \rot{AdaBoost-MLP} & \rot{KBA} & \rot{ACT} \\
		\hline
		TMD1 & 0.3712 & 0.1311 & 0.0505 & 0.0000 & 0.0000 & \textbf{0.5157} & 0.1267 & 0.0806 & 0.0000 & 0.0505 & 0.4827 \\
		TMD2 & \textbf{1.0000} & \textbf{1.0000} & \textbf{1.0000} & 1.0000 & \textbf{1.0000} & \textbf{1.0000} & \textbf{1.0000} & 0.0000 & 0.0000 & \textbf{1.0000} & \textbf{1.0000} \\
		TMD3 & \textbf{0.7195} & 0.4014 & 0.2136 & 0.0934 & 0.3847 & 0.4707 & 0.2668 & 0.2003 & 0.2287 & 0.4388 & 0.2594 \\
		TMD4 & \textbf{0.4457} & 0.1973 & 0.0000 & 0.2746 & 0.1439 & 0.1955 & 0.2808 & 0.2410 & 0.1849 & 0.1446 & 0.0000 \\
		TMD5 & 0.9940 & 0.8233 & 0.5811 & 0.0000 & 0.8471 & \textbf{1.0000} & 0.9018 & 0.9991 & 0.0129 & 0.6035 & 0.6012 \\
		TMD6 & 0.5159 & 0.4878 & 0.0000 & 0.3191 & 0.4878 & 0.6817 & 0.5691 & \textbf{0.7894} & 0.4273 & 0.4774 & 0.4878 \\
		TMD7 & \textbf{0.6885} & 0.6536 & 0.2000 & 0.5475 & 0.6536 & 0.6536 & 0.6273 & 0.6006 & 0.4532 & 0.6536 & 0.6536 \\
		TMD8 & 0.8106 & 0.7921 & 0.8268 & \textbf{0.8268} & 0.7462 & 0.7568 & 0.7921 & 0.7568 & 0.3585 & 0.7982 & 0.7921 \\
		TMD9 & \textbf{0.2669} & 0.0856 & 0.0444 & 0.0605 & 0.1969 & 0.2605 & 0.0000 & 0.2483 & 0.0993 & 0.0000 & 0.0000 \\
		TMD10 & 0.4585 & 0.4585 & 0.1633 & 0.4952 & 0.4585 & 0.4527 & 0.2000 & \textbf{0.4952} & 0.4588 & 0.0000 & 0.0000 \\
		TMD11 & 0.4983 & 0.3030 & 0.0000 & 0.3369 & 0.1973 & \textbf{0.5684} & 0.1908 & 0.3368 & 0.3295 & 0.3075 & 0.1955 \\
		TMD12 & 0.7738 & 0.4880 & 0.0000 & 0.4667 & 0.5120 & \textbf{0.9180} & 0.4880 & 0.4687 & 0.5258 & 0.2752 & 0.3416 \\
		TMD13 & \textbf{0.8054} & 0.3593 & 0.0000 & 0.0000 & 0.0000 & 0.7061 & 0.3646 & 0.2284 & 0.2076 & 0.4317 & 0.1937 \\
		TMD14 & 0.5044 & 0.2605 & 0.0000 & 0.0000 & 0.0472 & \textbf{0.6335} & 0.3697 & 0.0000 & 0.1573 & 0.1693 & 0.1578 \\
		TMD15 & \textbf{1.0000} & \textbf{1.0000} & 0.1993 & \textbf{1.0000} & 0.8162 & \textbf{1.0000} & 0.9375 & 0.7059 & 0.4472 & 0.5675 & 0.3414 \\
		TMD16 & \textbf{1.0000} & \textbf{1.0000} & 0.1414 & 0.2569 & 0.7928 & \textbf{1.0000} & 0.9693 & 0.8164 & 0.0192 & 0.6164 & 0.3628 \\
		TMD17 & 0.2095 & 0.2095 & 0.1631 & 0.0000 & 0.2000 & 0.2095 & 0.4341 & 0.2971 & 0.2998 & 0.2255 & \textbf{0.5846} \\
		TMD18 & \textbf{0.7593} & 0.4873 & 0.3911 & 0.0000 & 0.4003 & 0.5978 & 0.4651 & 0.2382 & 0.3970 & 0.4824 & 0.1754 \\
		TMD19 & 0.3933 & 0.2321 & 0.0000 & 0.0000 & 0.2104 & 0.2321 & 0.2942 & 0.1551 & 0.1452 & 0.1852 & \textbf{0.4753} \\
		TMD20 & 0.7035 & 0.6932 & 0.2914 & 0.6423 & 0.6932 & 0.7035 & \textbf{0.7365} & 0.3398 & 0.3835 & 0.6450 & 0.7361 \\
		TMD21 & \textbf{0.9854} & 0.9414 & 0.0000 & 0.0000 & 0.8018 & \textbf{0.9854} & 0.3842 & 0.5414 & 0.4402 & 0.6264 & 0.3854 \\
		TMD22 & 0.5882 & 0.6250 & 0.5872 & 0.6244 & 0.5789 & 0.5882 & 0.6977 & \textbf{0.7121} & 0.3436 & 0.6036 & 0.7095 \\
		TMD23 & 0.8310 & 0.4475 & 0.8310 & 0.4060 & \textbf{0.8437} & 0.7769 & 0.4088 & 0.2481 & 0.5108 & 0.2060 & 0.4065 \\
		TMD24 & \textbf{1.0000} & 0.8000 & 0.6691 & 0.4105 & 0.6000 & \textbf{1.0000} & 0.8564 & 0.3563 & 0.5104 & 0.4691 & 0.8000 \\
		TMD25 & 0.9028 & 0.7985 & 0.0230 & 0.8951 & 0.6300 & 0.7817 & 0.7109 & 0.6042 & 0.0002 & 0.6730 & \textbf{0.9559} \\
		TMD26 & 0.4330 & 0.4500 & 0.4584 & 0.3036 & \textbf{0.4610} & 0.4444 & 0.4000 & 0.3828 & 0.1081 & 0.3974 & 0.2717 \\
		TMD27 & 0.9922 & 0.9119 & 0.0000 & 0.0000 & 0.4600 & \textbf{0.9999} & 0.6333 & 0.0000 & 0.1508 & 0.4966 & 0.2881 \\
		TMD28 & \textbf{1.0000} & \textbf{1.0000} & 0.2000 & 0.0000 & 0.3420 & \textbf{1.0000} & 0.3774 & 0.0000 & 0.1292 & 0.7384 & 0.0000 \\
		TMD29 & \textbf{1.0000} & \textbf{1.0000} & 0.4116 & 0.9453 & 0.9414 & \textbf{1.0000} & 0.7715 & 0.7248 & 0.4106 & 0.7578 & \textbf{1.0000} \\
		TMD30 & \textbf{1.0000} & 0.8000 & 0.1526 & 0.6000 & 0.7526 & \textbf{1.0000} & 0.7794 & 0.0000 & 0.4472 & 0.8000 & \textbf{1.0000} \\
		TMD31 & 0.9681 & 0.9681 & 0.4898 & 0.8640 & 0.9285 & 0.9715 & 0.9681 & \textbf{1.0000} & 0.2927 & 0.9681 & 0.9681 \\
		TMD32 & \textbf{1.0000} & 1.0000 & 0.8877 & 0.6590 & 0.9771 & 0.9206 & 0.3825 & 0.9564 & 0.0777 & 0.6000 & 0.5642 \\
		TMD33 & 0.8766 & \textbf{0.9314} & 0.4812 & 0.3125 & 0.7766 & 0.8793 & 0.4803 & 0.8666 & 0.0996 & 0.4300 & 0.1300 \\
		TMD34 & \textbf{1.0000} & \textbf{1.0000} & 0.6875 & 0.8461 & \textbf{1.0000} & \textbf{1.0000} & \textbf{1.0000} & 0.9558 & 0.0000 & \textbf{1.0000} & \textbf{1.0000} \\
		TMD35 & \textbf{0.7642} & 0.1414 & 0.0000 & 0.0000 & 0.2083 & 0.4701 & 0.1067 & 0.4978 & 0.3013 & 0.0795 & 0.0000 \\
		TMD36 & \textbf{0.7237} & 0.5775 & 0.0661 & 0.3970 & 0.3935 & 0.6274 & 0.4840 & 0.6599 & 0.2802 & 0.4911 & 0.0000 \\
		TMD37 & \textbf{0.8000} & \textbf{0.8000} & 0.0000 & 0.1789 & \textbf{0.8000} & \textbf{0.8000} & 0.5586 & 0.6551 & 0.5477 & 0.4000 & 0.6904 \\
		TMD38 & \textbf{0.5296} & 0.0965 & 0.0965 & 0.0949 & 0.1438 & 0.2641 & 0.1022 & 0.1408 & 0.2550 & 0.0030 & 0.0000 \\
		TMD39 & \textbf{0.5479} & 0.4286 & 0.3185 & 0.1471 & 0.0000 & 0.4207 & 0.3313 & 0.4016 & 0.3616 & 0.4135 & 0.0000 \\
		TMD40 & 0.5559 & 0.1790 & 0.0000 & 0.0000 & 0.0000 & \textbf{0.6793} & 0.1491 & 0.0000 & 0.0000 & 0.0772 & 0.0000 \\
		TMD41 & \textbf{0.7246} & 0.3298 & 0.0000 & 0.5443 & 0.4402 & 0.1155 & 0.3298 & 0.6961 & 0.2110 & 0.4309 & 0.3298 \\
		TMD42 & \textbf{0.9738} & 0.6000 & 0.4000 & 0.0000 & 0.5859 & 0.8000 & 0.5968 & 0.3968 & 0.5449 & 0.6000 & 0.5996 \\
		\hline
		THD1 & 0.9740 & 0.9547 & 0.0000 & 0.9692 & 0.9558 & \textbf{0.9770} & 0.8951 & 0.9339 & 0.0487 & 0.7106 & 0.0000 \\
		THD2 & \textbf{1.0000} & 0.9797 & 0.0000 & 0.9631 & 0.9532 & 0.8165 & 0.9072 & 0.9464 & 0.0276 & 0.9797 & 0.0739 \\
		THD3 & \textbf{0.9961} & 0.9899 & 0.0000 & 0.8743 & 0.9395 & 0.9594 & 0.8434 & 0.9630 & 0.0022 & 0.9899 & 0.0000 \\
		THD4 & 0.0901 & 0.0060 & 0.0000 & 0.0508 & 0.0000 & \textbf{0.2493} & 0.0359 & 0.0000 & 0.1641 & 0.0000 & 0.1722 \\
		THD5 & \textbf{1.0000} & 0.4163 & 0.4163 & 0.8925 & 0.4163 & \textbf{1.0000} & 0.0000 & 0.5659 & 0.0000 & 0.1037 & 0.0000 \\
		THD6 & \textbf{1.0000} & 0.9414 & 0.9414 & 0.5431 & 0.9414 & \textbf{1.0000} & 0.0000 & 0.4839 & 0.4472 & 0.1414 & 0.0000 \\
		THD7 & 0.8853 & 0.7950 & 0.7375 & 0.7981 & 0.8721 & \textbf{0.9583} & 0.2119 & 0.8231 & 0.1867 & 0.6795 & 0.7234 \\
		THD8 & \textbf{1.0000} & 0.6691 & 0.6691 & 0.5066 & 0.7037 & \textbf{1.0000} & 0.0000 & 0.6125 & 0.0000 & 0.0691 & 0.0000 \\
		THD9 & 0.8920 & 0.8767 & 0.0000 & 0.0353 & 0.8431 & 0.8277 & 0.8848 & 0.0353 & 0.0401 & 0.8497 & \textbf{0.8953} \\
		THD10 & 0.7369 & 0.7351 & 0.0000 & 0.0005 & 0.9332 & 0.8151 & 0.7935 & 0.0005 & 0.0320 & 0.7146 & \textbf{0.9513} \\
		\hline
\end{tabular}
\end{center}
\end{table*}

\begin{table*}[!th]
\begin{center}
\scriptsize
\caption{ Results on multi-class datasets in terms of $Gmean$. }
\begin{tabular}{ c | c c c c c | c c c c c c } \hline
		\multirow{2}{*}{\parbox{0.3cm}{\vspace{0.9cm} \rot{Datasets}}} & \multicolumn{5}{c|}{OVO} & \multicolumn{5}{c}{OVA} \\ \cline{2-11}
		& \rot{KPBoost-SVM} & \rot{SVM} & \rot{AKS-$\chi^{2*}$} & \rot{RUSBoost$^{*}$} & \rot{AdaBoost-MLP$^{*}$} & \rot{KPBoost-SVM} & \rot{SVM} & \rot{AKS-$\chi^{2*}$} & \rot{RUSBoost$^{*}$} & \rot{AdaBoost-MLP$^{*}$} \\ \hline
		MMD1 & \textbf{1.0000} & 0.9773 & 0.5573 & 0.4709 & 0.9773 & 0.8527 & 0.7727 & 0.5573 & 0.4709 & \textbf{0.9773} \\
		MMD2 & 0.9347 & 0.9532 & 0.9263 & 0.9101 & \textbf{0.9923} & 0.9602 & 0.9475 & 0.9263 & 0.9101 & \textbf{0.9923} \\
		MMD3 & \textbf{0.0000} & \textbf{0.0000} & \textbf{0.0000} & \textbf{0.0000} & \textbf{0.0000} & \textbf{0.0000} & \textbf{0.0000} & \textbf{0.0000} & \textbf{0.0000} & \textbf{0.0000} \\
		MMD4 & \textbf{0.9388} & 0.9107 & 0.8712 & 0.8697 & 0.7725 & 0.8443 & 0.7128 & \textbf{0.8712} & 0.8697 & 0.7725 \\
		MMD5 & \textbf{0.8892} & 0.8305 & 0.5989 & 0.6645 & 0.8796 & 0.8486 & 0.8486 & 0.5989 & 0.6645 & \textbf{0.8796} \\
		MMD6 & \textbf{0.9232} & \textbf{0.9232} & 0.0000 & 0.0000 & 0.0000 & \textbf{0.9330} & 0.7712 & 0.0000 & 0.0000 & 0.0000 \\
		MMD7 & \textbf{0.9353} & \textbf{0.9353} & 0.9314 & 0.8788 & 0.8647 & \textbf{0.9687} & 0.9353 & 0.9314 & 0.8788 & 0.8647 \\
		MMD8 & \textbf{0.8292} & \textbf{0.8292} & \textbf{0.8292} & 0.7828 & 0.7860 & \textbf{0.8992} & 0.8292 & 0.8292 & 0.7828 & 0.7860 \\
		MMD9 & \textbf{1.0000} & 0.0000 & 0.0000 & 0.0000 & 0.0000 & \textbf{0.0000} & \textbf{0.0000} & \textbf{0.0000} & \textbf{0.0000} & \textbf{0.0000} \\
		MMD10 & 0.9583 & 0.6993 & 0.6875 & \textbf{0.9830} & 0.5346 & 0.8224 & 0.6247 & 0.6875 & \textbf{0.9830} & 0.5346 \\
		MMD11 & 0.9773 & 0.9773 & 0.9622 & \textbf{0.9846} & 0.9387 & \textbf{1.0000} & 0.9846 & 0.9622 & 0.9846 & 0.9387 \\
		MMD12 & 0.0000 & 0.0000 & \textbf{0.3870} & 0.0000 & 0.0000 & 0.3601 & 0.3649 & \textbf{0.3870} & 0.0000 & 0.0000 \\
\hline
		MHD1 & 0.9288 & 0.9201 & 0.9119 & 0.8770 & \textbf{0.9459} & 0.9282 & 0.9039 & 0.9119 & 0.8770 & \textbf{0.9459} \\
		MHD2 & \textbf{0.9340} & 0.8659 & 0.9334 & 0.9051 & 0.9107 & \textbf{0.9460} & 0.9200 & 0.9334 & 0.9051 & 0.9107 \\
		MHD3 & \textbf{0.0000} & \textbf{0.0000} & \textbf{0.0000} & \textbf{0.0000} & \textbf{0.0000} & \textbf{0.6405} & 0.0000 & 0.0000 & 0.0000 & 0.0000 \\
		MHD4 & 0.0000 & 0.0000 & \textbf{0.5658} & 0.0000 & 0.0000 & \textbf{0.7255} & 0.0000 & 0.5658 & 0.0000 & 0.0000 \\
		MHD5 & \textbf{0.0000} & \textbf{0.0000} & \textbf{0.0000} & \textbf{0.0000} & \textbf{0.0000} & \textbf{0.6971} & 0.6014 & 0.0000 & 0.0000 & 0.0000 \\
		MHD6 & \textbf{0.8595} & 0.7260 & 0.4517 & 0.6918 & 0.7736 & 0.8441 & \textbf{0.8462} & 0.4517 & 0.6918 & 0.7736 \\
		MHD7 & \textbf{0.9880} & 0.8298 & 0.0000 & 0.5253 & 0.9756 & \textbf{0.9880} & 0.9629 & 0.0000 & 0.5253 & 0.9756 \\
		MHD8 & \textbf{0.8660} & 0.0000 & 0.0000 & 0.0000 & 0.7154 & \textbf{0.9219} & 0.8784 & 0.0000 & 0.0000 & 0.7154 \\
		MHD9 & 0.7511 & 0.7462 & 0.4547 & 0.0000 & \textbf{0.7830} & 0.7556 & 0.6890 & 0.4547 & 0.0000 & \textbf{0.7830} \\ \hline
		\multicolumn{11}{l}{$*$ These algorithms do not have any OVO/OVA variants.}
\end{tabular}
\end{center}
\end{table*}

\begin{table*}[!th]
\begin{center}
\scriptsize
\caption{ Results on multi-class datasets in terms of $AUC$. }
\begin{tabular}{ c | c c c c c | c c c c c c } \hline
        \multirow{2}{*}{\parbox{0.3cm}{\vspace{0.9cm} \rot{Datasets}}} & \multicolumn{5}{c|}{OVO} & \multicolumn{5}{c}{OVA} \\ \cline{2-11}
		& \rot{KPBoost-SVM} & \rot{SVM} & \rot{AKS-$\chi^{2*}$} & \rot{RUSBoost$^{*}$} & \rot{AdaBoost-MLP$^{*}$} & \rot{KPBoost-SVM} & \rot{SVM} & \rot{AKS-$\chi^{2*}$} & \rot{RUSBoost$^{*}$} & \rot{AdaBoost-MLP$^{*}$} \\ \hline
        MMD1 & \textbf{1.0000} & 0.9833 & 0.7034 & 0.6299 & 0.9833 & 0.8931 & 0.8638 & 0.7034 & 0.6299 & \textbf{0.9833} \\
		MMD2 & 0.9667 & 0.9730 & 0.9590 & 0.9499 & \textbf{0.9955} & 0.9769 & 0.9702 & 0.9590 & 0.9499 & \textbf{0.9955} \\
		MMD3 & 0.8243 & 0.8243 & \textbf{0.8911} & 0.5591 & 0.8036 & \textbf{0.8946} & 0.8133 & 0.8911 & 0.5591 & 0.8036 \\
		MMD4 & \textbf{0.9663} & 0.9496 & 0.9251 & 0.9361 & 0.8845 & 0.9203 & 0.8695 & 0.9251 & \textbf{0.9361} & 0.8845 \\
		MMD5 & \textbf{0.9219} & 0.8785 & 0.7014 & 0.7674 & 0.9132 & 0.8941 & 0.8941 & 0.7014 & 0.7674 & \textbf{0.9132} \\
		MMD6 & \textbf{0.9516} & \textbf{0.9516} & 0.7916 & 0.6414 & 0.8246 & \textbf{0.9582} & 0.8661 & 0.7916 & 0.6414 & 0.8246 \\
		MMD7 & \textbf{0.9545} & \textbf{0.9545} & 0.9495 & 0.9101 & 0.9040 & \textbf{0.9773} & 0.9545 & 0.9495 & 0.9101 & 0.9040 \\
		MMD8 & \textbf{0.9100} & \textbf{0.9100} & \textbf{0.9100} & 0.8850 & 0.8875 & \textbf{0.9412} & 0.9100 & 0.9100 & 0.8850 & 0.8875 \\
		MMD9 & \textbf{1.0000} & 0.8750 & 0.8716 & 0.5708 & 0.7483 & 0.8091 & 0.8091 & \textbf{0.8716} & 0.5708 & 0.7483 \\
		MMD10 & 0.9700 & 0.8087 & 0.7854 & \textbf{0.9875} & 0.7708 & 0.8683 & 0.8063 & 0.7854 & \textbf{0.9875} & 0.7708 \\
		MMD11 & 0.9833 & 0.9833 & 0.9720 & \textbf{0.9886} & 0.9553 & \textbf{1.0000} & 0.9886 & 0.9720 & 0.9886 & 0.9553 \\
		MMD12 & 0.7858 & 0.7860 & 0.7255 & 0.6856 & \textbf{0.8156} & 0.6621 & 0.6758 & 0.7255 & 0.6856 & \textbf{0.8156} \\
		\hline
		MHD1 & 0.9574 & 0.9514 & 0.9457 & 0.9254 & \textbf{0.9670} & 0.9556 & 0.9411 & 0.9457 & 0.9254 & \textbf{0.9670} \\
		MHD2 & \textbf{0.9595} & 0.9190 & 0.9590 & 0.9416 & 0.9462 & \textbf{0.9665} & 0.9503 & 0.9590 & 0.9416 & 0.9462 \\
		MHD3 & \textbf{0.8042} & 0.7213 & 0.6718 & 0.6877 & 0.7695 & \textbf{0.8321} & 0.7338 & 0.6718 & 0.6877 & 0.7695 \\
		MHD4 & \textbf{0.8475} & 0.7772 & 0.7798 & 0.6837 & 0.6977 & \textbf{0.8641} & 0.7568 & 0.7798 & 0.6837 & 0.6977 \\
		MHD5 & \textbf{0.7897} & 0.7411 & 0.7505 & 0.6437 & 0.6939 & \textbf{0.8755} & 0.7981 & 0.7505 & 0.6437 & 0.6939 \\
		MHD6 & \textbf{0.9048} & 0.8259 & 0.7331 & 0.7763 & 0.8472 & \textbf{0.8919} & 0.8884 & 0.7331 & 0.7763 & 0.8472 \\
		MHD7 & \textbf{0.9911} & 0.8810 & 0.7411 & 0.6835 & 0.9821 & \textbf{0.9911} & 0.9732 & 0.7411 & 0.6835 & 0.9821 \\
		MHD8 & \textbf{0.9173} & 0.7540 & 0.7348 & 0.6799 & 0.8532 & \textbf{0.9511} & 0.9266 & 0.7348 & 0.6799 & 0.8532 \\
		MHD9 & 0.8361 & 0.8250 & 0.6222 & 0.5611 & \textbf{0.8500} & 0.8333 & 0.7889 & 0.6222 & 0.5611 & \textbf{0.8500} \\ \hline
		\multicolumn{11}{l}{$*$ These algorithms do not have any OVO/OVA variants.}
\end{tabular}
\end{center}
\end{table*}

\begin{table*}[!th]
\begin{center}
\scriptsize
\caption{ Results on multi-class datasets in terms of $GSDI$. }
\label{multiGSDI}
\begin{tabular}{ c | c c c c c | c c c c c c }
\hline
        \multirow{2}{*}{\parbox{0.3cm}{\vspace{0.9cm} \rot{Datasets}}} & \multicolumn{5}{c|}{OVO} & \multicolumn{5}{c}{OVA} \\ \cline{2-11}
		& \rot{KPBoost-SVM} & \rot{SVM} & \rot{AKS-$\chi^{2*}$} & \rot{RUSBoost$^{*}$} & \rot{AdaBoost-MLP$^{*}$} & \rot{KPBoost-SVM} & \rot{SVM} & \rot{AKS-$\chi^{2*}$} & \rot{RUSBoost$^{*}$} & \rot{AdaBoost-MLP$^{*}$} \\ \hline
        MMD1 & \textbf{1.0000} & 0.9773 & 0.5908 & 0.4128 & 0.9773 & 0.7806 & 0.7757 & 0.5908 & 0.4128 & \textbf{0.9773} \\
		MMD2 & 0.0032 & 0.7950 & 0.8637 & 0.8198 & \textbf{1.0000} & 0.8637 & 0.8637 & 0.8637 & 0.8198 & \textbf{1.0000} \\
		MMD3 & \textbf{0.0000} & \textbf{0.0000} & \textbf{0.0000} & \textbf{0.0000} & \textbf{0.0000} & \textbf{0.0000} & \textbf{0.0000} & \textbf{0.0000} & \textbf{0.0000} & \textbf{0.0000} \\
		MMD4 & \textbf{0.8134} & 0.8118 & 0.7329 & 0.3677 & 0.4728 & 0.3568 & 0.2301 & \textbf{0.7329} & 0.3677 & 0.4728 \\
		MMD5 & \textbf{0.7801} & 0.7555 & 0.6533 & 0.0001 & 0.0002 & \textbf{0.6748} & \textbf{0.6748} & 0.6533 & 0.0001 & 0.0002 \\
		MMD6 & \textbf{0.8732} & \textbf{0.8732} & 0.0000 & 0.0000 & 0.0000 & \textbf{0.8859} & 0.8292 & 0.0000 & 0.0000 & 0.0000 \\
		MMD7 & \textbf{0.9353} & \textbf{0.9353} & 0.8426 & 0.6037 & 0.8993 & \textbf{0.9687} & 0.9353 & 0.8426 & 0.6037 & 0.8993 \\
		MMD8 & \textbf{0.7454} & \textbf{0.7454} & 0.5499 & 0.5280 & 0.6853 & 0.5963 & 0.5499 & 0.5499 & 0.5280 & \textbf{0.6853} \\
		MMD9 & \textbf{1.0000} & 0.0000 & 0.0000 & 0.0000 & 0.0000 & \textbf{0.0000} & \textbf{0.0000} & \textbf{0.0000} & \textbf{0.0000} & \textbf{0.0000} \\
		MMD10 & 0.9291 & 0.6081 & 0.6599 & \textbf{0.9494} & 0.6335 & 0.8726 & 0.6693 & 0.6599 & \textbf{0.9494} & 0.6335 \\
		MMD11 & \textbf{1.0000} & \textbf{1.0000} & 0.9119 & 0.9119 & \textbf{1.0000} & \textbf{1.0000} & 0.9119 & 0.9119 & 0.9119 & \textbf{1.0000} \\
		MMD12 & 0.0000 & 0.0000 & \textbf{0.0019} & 0.0000 & 0.0000 & \textbf{0.0368} & 0.0015 & 0.0019 & 0.0000 & 0.0000 \\
		\hline
		MHD1 & 0.9148 & 0.9036 & 0.8796 & 0.8644 & \textbf{0.9559} & 0.9186 & 0.8897 & 0.8796 & 0.8644 & \textbf{0.9559} \\
		MHD2 & \textbf{0.8908} & 0.7646 & 0.8737 & 0.8227 & 0.8896 & \textbf{0.8949} & 0.8582 & 0.8737 & 0.8227 & 0.8896 \\
		MHD3 & \textbf{0.0000} & \textbf{0.0000} & \textbf{0.0000} & \textbf{0.0000} & \textbf{0.0000} & \textbf{0.1794} & 0.0000 & 0.0000 & 0.0000 & 0.0000 \\
		MHD4 & 0.0000 & 0.0000 & \textbf{0.1231} & 0.0000 & 0.0000 & \textbf{0.3206} & 0.0000 & 0.1231 & 0.0000 & 0.0000 \\
		MHD5 & \textbf{0.0000} & \textbf{0.0000} & \textbf{0.0000} & \textbf{0.0000} & \textbf{0.0000} & \textbf{0.3016} & 0.2050 & 0.0000 & 0.0000 & 0.0000 \\
		MHD6 & 0.6638 & 0.0532 & 0.5362 & 0.6279 & \textbf{0.9413} & 0.6808 & 0.7436 & 0.5362 & 0.6279 & \textbf{0.9413} \\
		MHD7 & \textbf{0.9527} & 0.6626 & 0.0000 & 0.0693 & 0.9434 & \textbf{0.9527} & 0.8405 & 0.0000 & 0.0693 & 0.9434 \\
		MHD8 & \textbf{0.8267} & 0.0000 & 0.0000 & 0.0000 & 0.3542 & 0.0048 & 0.0045 & 0.0000 & 0.0000 & \textbf{0.3542} \\
		MHD9 & \textbf{0.7076} & 0.6762 & 0.0029 & 0.0000 & 0.7012 & \textbf{0.7231} & 0.6222 & 0.0029 & 0.0000 & 0.7012 \\ \hline
		\multicolumn{11}{l}{$*$ These algorithms do not have any OVO/OVA variants.}
\end{tabular}
\end{center}
\end{table*}

\end{document}